\documentclass[journal]{IEEEtran}
\IEEEoverridecommandlockouts
\usepackage[numbers, sort&compress]{natbib}

\usepackage{amsmath,amssymb,amsfonts, bm}
\usepackage[ruled]{algorithm2e}
\usepackage{subfigure}
\usepackage{graphicx}
\usepackage{epstopdf}
\usepackage{textcomp}
\usepackage{xcolor}
\usepackage{authblk}
\usepackage[marginal]{footmisc}
\usepackage{multirow}
\usepackage{multicol}
\usepackage{makecell}
\def\BibTeX{{\rm B\kern-.05em{\sc i\kern-.025em b}\kern-.08em
    T\kern-.1667em\lower.7ex\hbox{E}\kern-.125emX}}

\title{TCGF: A unified tensorized consensus graph framework for multi-view representation learning}
\author{Xiangzhu Meng, Wei Wei$^*$, Qiang Liu, IEEE, Member, Shu Wu, IEEE, Senior Member, Liang Wang, IEEE, Fellow \thanks{Xiangzhu Meng and Wei Wei are corresponding authors (denoted by $\ast$)} \thanks{X. Meng is with the Center for Research on Intelligent Perception and Computing, Institute of Automation, Chinese Academy of Sciences, Beijing, China (xiangzhu.meng@cripac.ia.ac.cn)} \thanks{W. Wei is with Center for Energy, Environment \& Economy Research, School of Management, Zhengzhou University (weiwei123@zzu.edu.cn)} \thanks{Q. Liu, W. Shu and L. Wang are with Center for Research on Intelligent Perception and Computing Institute of Automation, Chinese Academy of Sciences, and School of Artificial Intelligence, University of Chinese Academy of Sciences (qiang.liu, shu.wu, liang.wang\}@nlpr.ia.ac.cn)} }

\begin{document}

\maketitle

\begin{abstract}
    Multi-view learning techniques have recently gained significant attention in the machine learning domain for their ability to leverage consistency and complementary information across multiple views.  However, there remains a lack of sufficient research on generalized multi-view frameworks that unify existing works into a scalable and robust learning framework, as most current works focus on specific styles of multi-view models. Additionally, most multi-view learning works rely heavily on specific-scale scenarios and fail to effectively comprehend multiple scales holistically. These limitations hinder the effective fusion of essential information from multiple views, resulting in poor generalization. To address these limitations, this paper proposes a universal multi-view representation learning framework named Tensorized Consensus Graph Framework (TCGF). Specifically, it first provides a unified framework for existing multi-view works to exploit the representations for individual view, which aims to be suitable for arbitrary assumptions and different-scales datasets. Then, stacks them into a tensor under alignment basics as a high-order representation, allowing for the smooth propagation of consistency and complementary information across all views. Moreover, TCGF proposes learning a consensus embedding shared by adaptively collaborating all views to uncover the essential structure of the multi-view data, which utilizes view-consensus grouping effect to regularize the view-consensus representation. To further facilitate related research, we provide a specific implementation of TCGF for large-scale datasets, which can be efficiently solved by applying the alternating optimization strategy. Experimental results conducted on seven different-scales datasets indicate the superiority of the proposed TCGF against existing state-of-the-art multi-view learning methods.
\end{abstract}


\section{Introduction}
With the rapid development of the big data era, more and more data can be obtained from different domains or described from various perspectives, which have gained extensive attention from researchers in recent years. For examples, the document could be translated as different versions via various languages \cite{bisson2012co}; an image could be represented by different visual descriptors \cite{douze2009evaluation, gao2008image} to reveal its color, texture, and shape information. Differing from single view, multi-view data are complementary to each other, which motivates the development of multi-view learning \cite{chao2021survey}. Considering the diversity of multiple views, it is essential for multi-view learning to exploit how to properly uncover rich information across views for improving practical performance.  

Multi-view learning methods have been extensively investigated in various applications, including classification \cite{zhang2019multi, feng2020multi}, clustering \cite{Zheng2018Binary, wang2019gmc}, and reidentification \cite{zhou2018aware, wang2020attribute}. This paper primarily focuses on the background of unsupervised multi-view learning without label information, such as clustering tasks. Graph-based multi-view methods \cite{Gu2017Multiple, Nie2017Auto, zhan2018graph, wang2019gmc, wang2019parameter, kang2020multi, liu2021multiview, meng2022unified} are prevalent since graphs can effectively represent data structures. The most representative group of graph-based multi-view methods aim to fuse multiple graphs into a common latent space shared by all views. For example, Multiple Kernel Learning (MKL) \cite{Gu2017Multiple, wang2019parameter} proposes a natural way to integrate different views by directly combining different views for learning a common representation. Unlike MKL, parameter-free multi-view learning methods \cite{Nie2017Auto} provide a self-weighting strategy to fuse multiple graph information without additional parameters. Furthermore, learning a shared graph among all views is an efficient way to integrate the diverse information within multi-view data, e.g., Graph-based Multi-view Clustering (GMC) \cite{wang2019gmc} and Multiview Latent Proximity Learning (MLPL) \cite{liu2021multiview}. To handle large-scale multi-view datasets, bipartite graph-based fast methods \cite{kang2020large, wang2021fast} have been proposed to obtain the consensus bipartite graph by linearly combining the bipartite graphs, which are adaptively learned from the corresponding views. Another group of representative multi-view learning methods is self-representation based multi-view learning, which learns a self-representation matrix to act as the affinity, such as Low-Rank Representation (LRR) \cite{liu2013robust, Tang2014Structure}, Sparse Subspace Clustering (SSC) \cite{elhamifar2013sparse, Jin2015A}, etc. Motivated by LRR and SSC, multi-view works \cite{Zhang2016Flexible, tang2018learning, xie2018unifying, meng2021multi} are further developed to learn affinity matrices based on the self-representation property. For example, MDcR \cite{Zhang2016Flexible} utilizes the complementarity information of multiple views based on the Hilbert Schmidt Independence Criterion (HSIC) as a regularization term to enhance the correlations across views and explores the correlations within each view jointly. The work \cite{tang2018learning} learns a shared affinity representation for multi-view subspace clustering by simultaneously considering the diversity regularization and a rank constraint.

Apart from the works on multi-view learning mentioned earlier, recent tensor-based multi-view learning methods \cite{xie2018unifying, zhang2015low, wen2021unified, chen2022low, jiang2022tensorial} have also played an important role in exploring diversity and complementary information across views. These methods attempt to capture high-order correlations among multiple views in tensor space by applying tensor nuclear norms \cite{xu2020low, gao2020enhanced, guo2022logarithmic} and tensor-Singular Value Decomposition (t-SVD) \cite{white2012convex}. For example, T-SVD-MSC \cite{xie2018unifying} models self-representation matrices of different views as tensors, which explores the consistency information of multiview data in tensor space. The work \cite{zhang2015low} aggregates multiple subspace representations into a third-order tensor, imposing a low-rank constraint by combining the nuclear norms of all matrices unfolded along each view. Furthermore, LTBPL \cite{chen2022low} utilized weighted tensor nuclear norm to recover the comprehensiveness in the low-rank constrained tensor stacked by multiple low-rank probability affinity matrices, and links consensus indicator graphs with view-specific representations carrying different adaptive confidences. The work \cite{wen2021unified} incorporates feature space based missing-view inferring with low-rank tensor constraint to recover the missing views and explore the full information of the recovered views and available views. IMVTSC-MVI \cite{jiang2022tensorial} utilizes tensorial modeling to capture both the pairwise correlations between samples and the higher-order correlations between features, resulting in a more accurate and robust multi-view clustering method.

\subsection{Motivations}

Despite the significant progress made by current multi-view learning methods in practical applications, there are still several limitations that require further attention. One such limitation is the heavy dependence of some methods on pre-defined adjacency matrices, which hinders their flexibility and limits their applicability to various multi-view scenarios. For example, many graph-based methods rely heavily on predefined similarity matrices of different views. However, due to the complexity and unknown prior information about the geometric structure of views, it remains an open problem to manually construct a suitable similarity matrix, often resulting in suboptimal performance. Meanwhile, we also observe that most multi-view learning works rely heavily on specific-scale scenarios, which prevents them from comprehensively understanding different-scales multi-view scenarios. For instance, bipartite graph-based multi-view methods are often used to efficiently process large-scale multi-view datasets, but their performance is not as promising when applied to normal-scale scenarios.  Similarly, graph-based or tensor-based multi-view learning methods also encounter difficulties in dealing with large-scale datasets, owing to their computational complexity of squared or cubic with the data size.

Given the success of multi-view learning methods in achieving impressive performance, it is evident that views fusion and tensor construction play important and essential roles in promoting correlation and consistency among multiple views. Therefore, how to propose a more flexible and robust multi-view learning method based on these mentioned earlier works is essential yet full of challenges problem. To tackle these challenges, this paper attempts to propose a unified multi-view learning framework that can transform a wide range of existing multi-view learning approaches into a unified formulation, thereby enhancing its flexibility and suitability for various multi-view scenarios.

\subsection{Contributions}
To simultaneously address the above limitations, this paper proposes a novel multi-view representation learning framework named Tensorized Consensus Graph Framework (TCGF). It first provides a unified framework for existing multi-view works to learn the view-specific basics and representations, and then stacks representations of all views into a tensor as a high-order representation. Then, TCGF utilizes weighted tensor singular value decomposition (t-SVD) based tensor rank minimization to exploit the complementary information among different views. Different existing tensor-based muli-view learning works, it's scalable to construct the weighted t-SVD operator based on the axis for basics or instances, enhancing its potential ability to uncover complementary information. Moreover, the view-consensus grouping effect is formulated between the view-specific representations and consensus representation to capture the consensus information across different views, which enforces the multi-view smooth regularization on shared space. Consequently, TCGF not only leverages most existing embedding works into a unified formulation but simultaneously considers the diversity, complementary and consensus information across multiple views. Notably, we additionally observe that it's feasible to select the appropriate embedding manner as well as its basics to handle different-scale multi-view datasets. To comprehensively validate the effectiveness of TCGF, we conduct massive experiments on seven multi-view datasets. Experimental results demonstrate that the proposed TCGF can outperform  the current state-of-the-art multi-view learning methods in most situations. The major contributions in this paper can be summarized as follows:

\begin{itemize}
\item We propose a novel multi-view representation learning framework named TCGF to learn the shared embedding, which serves as a unified framework for existing multi-view works.

\item The axis-free weighted t-SVD operator is designed to exploit the complementary information among different views, which further extends the applications of tensor rank minimization.

\item We propose the view-consensus grouping effect to regularize the view-consensus representation, which enables the discovery of essential structure information within the multi-view data.

\item The experimental results conducted on seven different-scale datasets demonstrate that TCGF is not only capable of maintaining or surpassing the performance of other state-of-the-art multi-view methods, but also adaptable to datasets with different scales.

\end{itemize}

\section{Related work}\label{related_works}
Existing multi-view methods can be divided into two categories according to the means of calculating the affinity representation, i.e., graph-based and self-representation-based multi-view models. Besides, tensor-based multi-view methods also play an important role in exploring complemenary information across views.

\subsection{Graph-based Multi-view Learning}
The graph-based learning framework involves learning an affinity matrix $\bm{S}^v$ that encodes the similarity between different samples in the $v$th view. This affinity matrix $\bm{S}^v$ is learned by minimizing the distance between samples in the latent space, which can be formulated as
\begin{equation}
    \mathop {\min }\limits_{{\bm{S}^v} \in \mathbf{\bm{C}}} \sum\limits_{i= 1}^N {\sum\limits_{j=1}^N} {d(\bm{x}^v_i, \bm{x}^v_j)\bm{S}_{i,j}^v} + \lambda \bm\Omega(\bm{S}^v),
\end{equation}
where $d(\bm{x}^v_i, \bm{x}^v_j)$ denotes the distance between two samples, which can be calculated by $L_1$-norm, Euclidean distance, Mahalanobis distance, etc. $\mathbf{\bm{C}}$ and $\bm\Omega(\bm{S}^v)$ stands by the constraint and normalization terms on the affinity matrix $\bm{S}^v$, respectively.
Based on the view-specific affinity matrix $\bm{S}^v$, graph-based multi-view methods aim to learn an intrinsic representation, capturing both consistent and complementary information among multiple views. The most representative group of multi-view methods \cite{ma2017multi, Gu2017Multiple,
Nie2017Auto, zhan2018graph, wang2019gmc, wang2019study, wang2019parameter, kang2020multi, liu2021multiview} aim to fuse multiple features or graphs into one common latent space shared by all views. Multiple Kernel Learning (MKL) \cite{Gu2017Multiple, wang2019parameter} is also a natural way to integrate different views based on the direct combination of different views and learns a common low-dimensional representation. Different from MKL, parameter-free multi-view learning methods \cite{Nie2017Auto} provide a self-weighting strategy to fuse multiple graph information without additional parameters. Besides, learning a shared graph among all views is also an efficient manner to integrate the diversity information within multi-view data, e.g. Graph-based Multi-view Clustering (GMC) \cite{wang2019gmc} and Multiview Latent Proximity Learning (MLPL) \cite{liu2021multiview}. Due to the squared computational complexity of graph-based works, these methods might be inefficient in dealing with large-scale multi-view datasets. For this reason, bipartite graph-based multi-view methods \cite{kang2020large, wang2021fast, shu2022self} have aroused widespread research interest to reduce both the computational complexity and storage complexity, where the bipartite graph can well present the relationship between $N$ samples and $K$ ($K \ll N$) anchors. It is worth noting that the performance of the aforementioned graph-based methods heavily depends on the quality of the predefined view-specific affinity matrix $\bm{S}^v$. However, it is still an open problem to manually construct a suitable similarity matrix for each view due to the complex and unknown geometric structure of multi-view data, which limits their applicability.

\subsection{Self-representation based Multi-view Learning}
Another graph-based ones are using the so-called self-expressiveness property, which are developed to learn affinity matrices based on the self-representation property. Specifically, the self-representation method is an important subspace learning technology based on generating a large number of representative samples, which can be formulated as
\begin{equation}
    \mathop {\min }\limits_{{\bm{S}^v} \in \mathbf{\bm{C}}} {\left\| \bm{X}^v - \bm{X}^v \bm{S}^v\right\|}_2 + \lambda \bm\Omega(\bm{S}^v),
\end{equation}
where the normalization term $\bm\Omega(\bm{S}^v)$ on the affinity matrix $\bm{S}^v$ is the core component. For example, the work \cite{liu2013robust} adopts the low-rank constraint as normalization term $\bm\Omega(\bm{S}^v)$, which aims to approximately recover the row space with theoretical guarantees to remove arbitrary sparse errors. Motivated by LRR \cite{liu2013robust, Tang2014Structure} and SSC \cite{elhamifar2013sparse, Jin2015A}, multi-view works \cite{Zhang2016Flexible, tang2018learning, xie2018unifying} are further developed to learn affinity matrices based on the self-representation property. For example, MDcR \cite{Zhang2016Flexible} utilizes the complementarity information of multiple views based on the Hilbert Schmidt Independence Criterion (HSIC) as a regularization term to enhance the correlations across views and explores the correlations within each view jointly. The work \cite{tang2018learning} learns a shared affinity representation for multi-view subspace clustering by simultaneously considering the diversity regularization and a rank constraint.
Even though the above self-representation multi-view works obtain impressive performance and efficiency, there still exist the following limitation in most works. Most of them aim to study a common representation or the pairwise correlations between views, leading to the loss of comprehensiveness and deeper higher-order correlations among multi-view data, and hence miss important underlying semantic information.


\begin{figure*}[htbp]
  \centering
  \includegraphics[width = \textwidth]{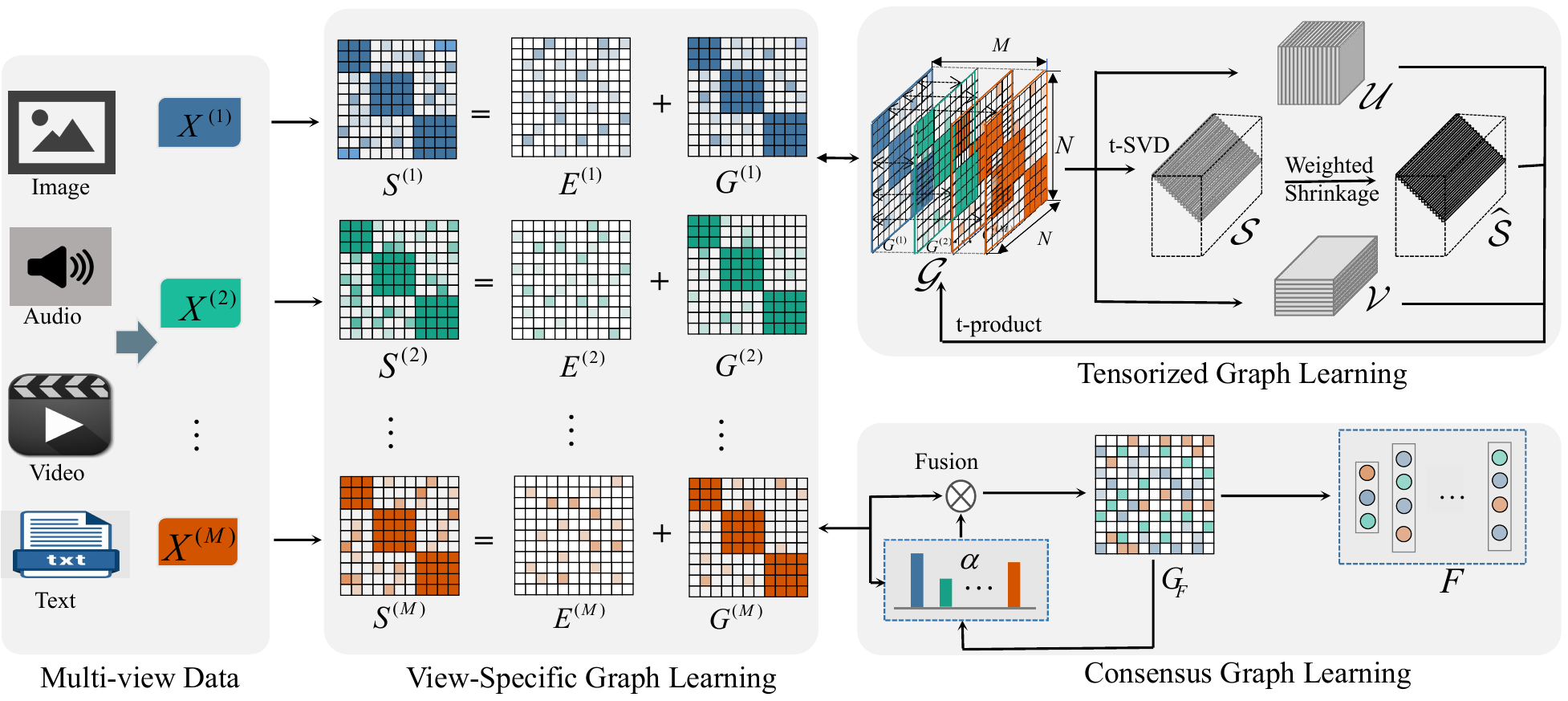}
   \caption{Flowchart of the proposed TCGF. Given a collection of samples with $M$ views, e.g., $\{\bm{X}^{(1)}, \bm{X}^{(2)}, \ldots, \bm{X}^{(M)}\}$. TCGF first explores the view-specific graph $\bm{G}^{v}$ by removing the disturbance of the noise error $\bm{E}^{(v)}$ in the initialized graph $\bm{S}^{(v)}$. After stacking view-specific graphs $\{\bm{G}^{(1)}, \bm{G}^{(2)}, \ldots, \bm{G}^{(M)}\}$ into one tensor $\bm{\mathcal{G}}$, the tensor $\bm{\mathcal{G}}$ can be updated by utilizing the t-SVD based weighting tensor multi-rank minimization. Based on the graph agreement term, the consensus graph $\bm{G}_F$ can be obtained by fusing the view-specific graphs $\{\bm{G}^{(1)}, \bm{G}^{(2)}, \ldots, \bm{G}^{(M)}\}$ with adaptively allocated weights $\bm\alpha$. Taking the view-specific graph learning, tensorized graph learning and consensus graph learning into one whole framework, we can obtain the final latent embedding $\bm{F}$ shared by all views. }
      \label{flow-chart}
\end{figure*}

\subsection{Tensor-based Multi-view learning}
To capture the high-order correlations among multiple views, tensor-based multi-view learning methods \cite{xie2018unifying, zhang2015low, wen2021unified, chen2022low, jiang2022tensorial} have been also developed in recent years, which play an important role in effectively exploring the comprehensive information among multiple views. The core idea of tensor-based multi-view learning methods is to model high-order correlations across views in tensor space by the low-rank constraint based on different tensor nuclear norms \cite{xu2020low, gao2020enhanced, guo2022logarithmic}. For example, T-SVD-MSC \cite{xie2018unifying} provides a tensor low-rank constraint on the stacked subspace representation matrices to capture the high-order complementary information among multiple views via introducing the tensor singular value decomposition. The work \cite{zhang2015low} aggregates multiple subspace representations into a third-order tensor, imposing a low-rank constraint by combining the nuclear norms of all matrices unfolded along each view. The work \cite{wen2021unified} studied the t-SVD based weighted tensor nuclear norm minimization to shrink different matrix singular values with the corresponding confidence.  Moreover, TMvC\cite{jiang2022tensorial} proposes the high-order graph to uncover the essential information stored in multiple views, which adopts the low-rank constraint of horizontal and vertical directions to better uncover the inter-view and inter-class correlations between multi-view data. The work \cite{wen2021unified} incorporates feature space-based missing-view inferring with low-rank tensor constraint to recover the missing views and explore the full information of the recovered views and available views. IMVTSC-MVI \cite{jiang2022tensorial} utilizes tensorial modeling to capture both the pairwise correlations between samples and the higher-order correlations between features, resulting in a more accurate and robust multi-view clustering method. Although these tensor-based approaches have achieved promising effects, most of these methods mainly focus on graph-based settings, resulting in cubic computational complexity. Meanwhile, they usually fail to simultaneously consider the inter-view and intra-view relationships among samples.

\section{The Proposed Method}
In this section, we first introduce the main notations and definitions used in this paper. Then, we provide the construction process of the proposed TCGF in detail. Correspondingly, we provide a typical implement for our proposed framework for large-scale datasets. For clarity, the flow chart of TCGF is shown in Fig. \ref{flow-chart}.

\subsection{Notations and Problem Definition}
We use bold calligraphy letters for tensors, e.g., $\bm{\bm{\mathcal{A}}} \in \mathbb{R}^{n_1 \times n_2 \times n_3}$, bold upper case letters for matrices, e.g., $\bm{A} \in \mathbb{R}^{n_1 \times n_2}$. $\bm{A}(i,j)$ denotes the $(i,j)$th entry of the matrix $\bm{A}$. $\bm{A}(i,:)$ denotes the $i$th row of the matrix $\bm{A}$. $\bm{\mathcal{A}}^T \in \mathbb{R}^{n_2 \times n_1 \times n_3}$ denotes the transpose tensor of tensor $\bm{\mathcal{A}}$. The fast Fourier transformation (FFT) along the third axis of a tensor $\bm{\mathcal{A}}$ and its inverse operation are $\bm{\mathcal{A}}_f=fft( \bm{\mathcal{A}},  \left[ \right], 3)$ and $\bm{\mathcal{A}}=ifft(\bm{\mathcal{A}}_f, \left[ \right], 3)$, respectively. The block vectorizing and its inverse operation of $\bm{\mathcal{A}}$ are $bvec(\bm{\mathcal{A}})=[\bm{\mathcal{A}}^{(1)}, \bm{\mathcal{A}}^{(2)}, \cdots, \bm{\mathcal{A}}^{(n_3)}] \in \mathbb{R}^{n_1 n_2 \times n_3}$ and $fold(bvec(\bm{\mathcal{A}}))=\bm{\mathcal{A}}$. $bcirc(\bm{\mathcal{A}}) \in \mathbb{R}^{n_1 n_2 \times n_1 n_3}$ denotes block circulant matrix of tensor $\bm{\mathcal{A}}$. The $L_1$ norm of matrix $\bm{A}$ is denoted as $\left\| \bm{A} \right\|_1$. $tr(\bm{A})$ denotes the trace of matrix $\bm{A}$. $\bm{\bm{\mathcal{I}}}$ denotes the $ n_1 \times n_2 \times n_3 $ identify tensor. $\bm{I}$ denotes the identity matrix. $\bm{1}$ denotes a vector whose elements are equal to 1.

The tensor singular value decomposition operation (t-SVD) and tensor nuclear norm are defined as follows.

\textbf{Definition 1} (t-product): For two tensors $\bm{\mathcal{A}} \in \mathbb{R}^{n_1 \times n_2 \times n_3}$ and $\bm{\mathcal{A}} \in \mathbb{R}^{n_2 \times n_4 \times n_3}$, the t-product $\bm{\mathcal{A}} \ast \bm{\mathcal{B}} =fold(bcirc(\bm{\mathcal{A}})bvec(\bm{\mathcal{B}}))$ is the $n_1 \times n_4 \times n_3$ tensor.

\textbf{Definition 2} (t-SVD): The t-SVD of tensor $\bm{\mathcal{A}} \in \mathbb{R}^{n_1 \times n_2 \times n_3}$ is defined as $\bm{\mathcal{A}} = \bm{\mathcal{U}} \ast \bm{\mathcal{S}} \ast \bm{\mathcal{V}}^T$, where $\bm{\mathcal{U}} \in \mathbb{R}^{n_1 \times n_1 \times n_3} $ and $\bm{\mathcal{V}}  \in \mathbb{R}^{n_2 \times n_2 \times n_3} $ are two orthogonal tensors, i.e., $\bm{\mathcal{U}} \ast \bm{\mathcal{U}}^T=\bm{\mathcal{U}}^T \ast \bm{\mathcal{U}}=\bm{\mathcal{I}}$ and $\bm{\mathcal{V}} \ast \bm{\mathcal{V}}^T=\bm{\mathcal{V}}^T \ast \bm{\mathcal{V}}=\bm{\mathcal{I}}$. $\bm{\mathcal{S}} \in \mathbb{R}^{n_1 \times n_2 \times n_3}$ is a f-diagonal tensor, whose entire frontal slices are diagonal matrices.

\textbf{Definition 3} (Weighted t-SVD based tensor nuclear norm): For a tensor $\bm{\mathcal{A}}$, the weighted t-SVD based tensor nuclear norm $\bm{\mathcal{A}}_{\bm{\omega}, \ast}$ is defined as by the weighted sum of singular values of all the frontal slices of  $\bm{\mathcal{A}}_f$, i.e., $\bm{\mathcal{A}}_{\bm{\omega},\ast}=\sum_{i=1}^{min\{n_1, n_2\}}\sum_{j=1}^{n_3}\bm{\omega}_i|\bm{\mathcal{S}}_f^{(j)}(i, i)|$. $\bm{\omega}$ is the weighted coefficients.

\subsection{Model Formulation}
\subsubsection{View-specific Graph Learning}
Given a multi-view dataset consisting of $M$ views, the data in the $v$th view ($1 \le v \le M$) can be denoted as $\bm{X}^{(v)} = \{\bm{x}_1^{(v)}, \bm{x}_2^{(v)}, \ldots, \bm{x}_N^{(v)}\}$, in which $N$ is the number of samples. $\bm{S}^{(v)} \in \mathbb{R}^{N \times N}$ denotes the initialized graph in the $v$th view, which reflect the relationships among samples. For the construction of specific-view graph $\bm{S}^{(v)}$, there are different manners to be flexibility adopted, such as similarity and self-representation graphs. We use $\bm\Omega(\bm{S}^{(v)})$ to generally represent the objective function that constructs the original graph $\bm{S}^{(v)}$. Notably, there are usually noise and redundant information in the original multi-view data, which might result in the errors $\bm{E}^{(v)}$ for view-specific graph $\bm{S}^{(v)}$. To eliminate the efforts of error, the basic model of graph learning can be formulated as follows:
\begin{equation}
\begin{split}
& \mathop {\min }\limits_{\bm{G}^{(v)}, \bm{E}^{(v)} } \sum_{v=1}^M {\bm\Omega(\bm{S}^{(v)})} + \lambda_E \sum_{v=1}^M {\left\|\bm{E}^{(v)}\right\|_1}, \\
&  s.t. \bm{S}^{(v)}=\bm{G}^{(v)}+\bm{E}^{(v)}, \bm{G}^{(v)}\ge 0, \bm{G}^{{(v)}^T}\bm{1}=\bm{1}, \\
\end{split}
\end{equation}
where the constraint $\bm{G}^{{(v)}^T} \bm{1}=\bm{1}$ guarantees that the sum of graph weights between each sample and other all samples 1. $\lambda_E$ is the trade-off parameter.

\subsubsection{Tensorized Graph Learning}
Inspired by tensor nuclear norm, which well exploits complementary information and spatial structure embedded in tensor, we utilize the weighted t-SVD based tensor nuclear norm to investigate the high-order correlations among multiple views. Thus, we stack the affinity matrices of all views into a tensor $\bm{\mathcal{G}} \in \mathbb{R}^{N \times N \times M}$. To better investigate the correlations and largely reduce the computational complexity, we further rotate $\bm{\mathcal{G}}$ into a $N \times M \times N$ tensor, where $\bm{\mathcal{G}}(:, v, :)=\bm{G}^{(v)}$. After rotation,  the dimension of the weighted coefficient $\bm{\omega}$ decreases from $N$ to $M$ ($M \ll N$), in which finetuned parameters $\bm{\omega}$ correlates with the view number. Considering the influence of noise on $\bm{\mathcal{G}}$, we learn a low-rank tensor $\bm{\mathcal{Z}}$ to approximate $\bm{\mathcal{G}}$ as follows:
\begin{equation}
\begin{split}
&\mathop {\min}\limits_{ \bm{\mathcal{Z}}} \quad \bm{\mathcal{Z}}_{\bm{\omega}, \ast}, \\
&s.t. \quad \bm{\mathcal{Z}} = \bm{\mathcal{G}}, \  \bm{\mathcal{G}}=\Phi\left(\bm{G}^{(1)}, \cdots, \bm{G}^{(M)}\right),\\
\end{split}
\end{equation}
where $\Phi(\cdot)$ merges graphs of multiple views into a tensor and then rotates along the third axis.

\subsubsection{Consensus Embedding Learning}
As multi-view features are extracted from the same objects, different graphs $\{\bm{G}^{(v)}\}$ should contain some similar information. Thus, it's essential to capture both consistent and complementary information among multiple views. For this reason, we propose the shared embedding $\bm{F} \in \mathbb{R}^{N \times d}$ to uncover the rich information in each view.  To preserve the locality in the learned graph $\bm{G}^{(v)}$ in each view, we exploit the subspace-wise grouping effect \cite{hu2014smooth} in the learned graph $\bm{G}^{(v)}$ by means of a unified framework.

\textbf{Definition 4} (Subspace-wise Grouping Effect): Given a set of d-dimensional data points $\bm{X}=[\bm{x}_1, \bm{x}_2, \cdots, \bm{x}_n] \in \mathcal{R}^{d \times n}$, a self-representation matrix $\bm{Z}=[\bm{z}_1, \bm{z}_2, \cdots, \bm{z}_n] \in \mathcal{R}^{n \times n}$ has the grouping effect if $ \left|\bm{x}_i - \bm{x}_j \right|^2 \rightarrow 0 $ then $ \left| \bm{z}_i - \bm{z}_j \right|^2 \rightarrow 0 $.

According to the above definition, we can formulate the view-consensus term as $tr\left(\bm{F}^T \bm{G}^{(v)} \bm{F} \right)$. To further improve its flexibility, we propose the consensus graph based on shared embedding  $\bm{F}$. Meanwhile, considering different views have different contributions in learning $\bm{G}_{\bm{F}}$, we adaptively assign weight $\bm{\alpha}_v \leq 0$ for the $w$th view. The above considerations can be formed as follows:
\begin{equation}
\begin{split}
&\mathop {\max}\limits_{ \bm{F} } \sum_{v=1}^M { {(\bm{\alpha}^{(v)})}^{\gamma} tr\left(\bm{F}^T \bm{F} \bm{G}^{(v)} {\bm{D}^{(v)}}^{-1}\right)}, \\
&s.t. \quad \sum_{v=1}^M \bm{\alpha}^{(v)}=1,  \bm{\alpha}^{(v)} \ge 0,
     \bm{F}^T \bm{F} = \bm{I},\\
\end{split}
\end{equation}
where $\gamma$ is a hyper-parameter. $\bm{D}^{(v)}$ is a diagonal matrix whose diagonal elements are $\bm{D}^{(v)}(i, i)=\sum_{i=j}^N\bm{D}^{(v)}(j, i)$. $tr\left(\bm{G}_{\bm{F}} \bm{G}^{(v)} {\bm{D}^{(v)}}^{-1}\right)$ can be seen as the graph agreement term to measure the consistence between the consensus graph $\bm{G}_{\bm{F}}$ and the normalized graph $\bm{G}^{(v)} {\bm{D}^{(v)}}^{-1}$. Notably, the construction manner for graph $\bm{G}_{\bm{F}}$ is also to be flexibly chosen, such as linear kernel function $\bm{G}_{\bm{F}}=\bm{F}\bm{F}^T$.


\subsubsection{Overall Framework of TCGF}
In the above subsections, we discuss how to learn the intrinsic, consistent and complement information in the multi-view data. To this end, we seamlessly couple the above learning processes, and the overall objective function is formulated as follows:

\begin{equation}\label{eq:total_loss}
\begin{split}
&\mathop {\min}_{\bm{F}, \bm{\mathcal{Z}}, \bm{S}^{(v)}, \bm{G}^{(v)},  \bm{E}^{(v)}, \bm{\alpha}^{(v)}} \underbrace{\sum_{v=1}^M {\bm\Omega(\bm{S}^{(v)})} + \lambda_E \sum_{v=1}^M {\left\|\bm{E}^{(v)}\right\|_1}}_{View-specific \ Term} \\
&  - \underbrace{\lambda_C \sum_{v=1}^M { {(\bm{\alpha}^{(v)})}^{\gamma} tr(\bm{G}_{\bm{F}} \bm{G}^{(v)} {\bm{D}^{(v)}}^{-1})}}_{Consensus \ Graph \ Learning} + \underbrace{\lambda_R \bm{\mathcal{Z}}_{\bm{\omega}, \ast}}_{Tensorized \ Term},\\
& s.t. \quad \bm{S}^{(v)}=\bm{G}^{(v)}+\bm{E}^{(v)}, \bm{G}^{(v)}\ge 0, \bm{G}^T\bm{1}=\bm{1},\\
&  \qquad \sum_{v=1}^M \bm{\alpha}^{(v)}=1,  \bm{\alpha}^{(v)} \ge 0,
     \bm{F}^T \bm{F} = \bm{I}, \\
& \qquad \bm{\mathcal{Z}} = \bm{\mathcal{G}}, \bm{\mathcal{G}}=\Phi(\bm{G}^{(1)}, \cdots, \bm{G}^{(M)}), \\
\end{split}
\end{equation}
where $\lambda_E$ and $\lambda_T$ are trade-off parameters. Observed from the model in Eq. (\ref{eq:total_loss}), the shared embedding $\bm{F}$ and view-specific graph $\bm{G}^{(v)}$ that is constrained by the low-rank tensor $\bm{\mathcal{G}}$ can be simultaneously learned in a unified framework. The first aspect maintains consensus information among different views to obtain the shared embedding, via fusing multiple graph agreement terms with different adaptive weights. The second aspect is to eliminate the efforts of error, which can learn more robust graph $\bm{G}^{(v)}$. The final aspect depicts the low-rank property and higher-order correlations of affinity tensor $\bm{\mathcal{G}}$ to exploit complementary information among views.

\subsection{Specific Implement for Large-scale Datasets}
Due to the computational complexity of the aforementioned methods being squared or cubic with the data size, thus they are inefficient in handling large-scale datasets. To solve this issue, we attempt to provide a specific implement to extend the proposed TCGF into the scenario of large-scale datasets.

Inspired by bipartite graph, we attempt to reduce the scale of initialized graph $\bm{S}^{(v)}$ by randomly selecting the subset of samples as anchors. Then, we construct the bipartite graph $\bm{B}^{(v)}$ between samples and anchors to substitute the whole graph $\bm{S}^{(v)}$. However, directly using those anchors is difficult to cover the entire data point clouds of data and characterize the intrinsic structure of data. To solve this issue, we propose a novel anchors selection scheme based on the information volume, which is simple and efficient. We combine all views into one view, and then execute the SVD decomposition to obtain the singular values for all samples. In this way, we select the $K$ most representative samples according to their singular values. Besides, we also find that K-means is also utilized to find those samples with high information volume in some situations. After that, we can construct the bipartite graph $\bm{B}^{(v)} \in \mathcal{R}^{N \times K}$ to re-initialize the view-specific graph. Moreover, to further control the computational cost, we generate similarity-induced graph to construct the fixed bipartite graph $\bm{B}^{(v)}$. According to the above considerations, we can readily extend the proposed TCGF into the scenario of large-scale datasets, which can be formulated as follows:
\begin{equation}
\begin{split}
&\mathop {\min}_{\bm{\Theta}}  -\sum_{v=1}^M { {(\bm{\alpha}^{(v)})}^{\gamma} tr(\bm{G}_{\bm{F}} \bm{G}^{(v)} {\bm{D}^{(v)}}^{-1})} + \lambda_E \sum_{v=1}^M {\left\|\bm{E}^{(v)}\right\|_1}
\\ &  + \lambda_R \bm{\mathcal{Z}}_{\bm\omega, \ast},\\
& s.t. \quad \bm{B}^{(v)}=\bm{G}^{(v)}+\bm{E}^{(v)}, \bm{G}^{(v)}\ge 0, \bm{G}^{{(v)}^T}\bm{1}=\bm{1},\\
&  \qquad \sum_{v=1}^M \bm{\alpha}^{(v)}=1,  \bm{\alpha}^{(v)} \ge 0,
     [\bm{F}_S; \bm{F}_A]^T [\bm{F}_S; \bm{F}_A] = \bm{I}, \\
& \qquad \bm{\mathcal{Z}} = \bm{\mathcal{G}}, \bm{\mathcal{G}}=\Phi(\bm{G}^{(1)}, \cdots, \bm{G}^{(M)}), \\
\end{split}
\end{equation}
where $\bm{\Theta}=\{\bm{F}_S, \bm{F}_A,\bm{\mathcal{Z}}, \bm{G}^{(v)},  \bm{E}^{(v)}, \bm{\alpha}^{(v)} \}$ denotes the set of solved variables. $\bm{F}_S \in \mathcal{R}^{N \times d}$ and $\bm{F}_A \in \mathcal{R}^{K \times d}$ represent the shared embedding of samples and anchors, respectively.

\subsubsection{Optimization}
Inspired by the augmented Lagrange multiplier method, the corresponding augmented Lagrangian function of the Eq. (\ref{eq:total_loss}) can be formulated as follows:
\begin{equation}\label{eq:lang}
\scriptsize
\begin{split}
    & \bm{\mathcal{L}}(\bm{F}_S, \bm{F}_A, \bm{\mathcal{Z}}, \bm{G}^{(v)}, \bm{E}^{(v)}, \bm{\alpha}^{(v)}) = \\
    & -\sum_{v=1}^M { {(\bm{\alpha}^{(v)})}^{\gamma} tr(\bm{F}^T \bm{G}^{(v)} {\bm{D}^{(v)}}^{-1}\bm{F})}  + \lambda_E \sum_{v=1}^M {\left\|\bm{E}^{(v)}\right\|_1} + \lambda_R \bm{\mathcal{Z}}_{\bm\omega, \ast}\\
    &+\sum_{v=1}^M  \left(	\left \langle\bm{Y}^{(v)}, \bm{S}^{(v)}-\bm{G}^{(v)}-\bm{E}^{(v)}	\right \rangle + \frac{\mu}{2}\left\| \bm{S}^{(v)}-\bm{G}^{(v)}-\bm{E}^{(v)} \right\|_F^2 \right ) \\
    & + \left \langle \bm{\mathcal{Y}}, \bm{\mathcal{G}} - \bm{\mathcal{Z}} \right \rangle + \frac{\rho}{2}\left\| \bm{\mathcal{G}} - \bm{\mathcal{Z}} \right\|_F^2,\\
\end{split}
\end{equation}
where $\bm{Y}^{(v)}$ and $\bm{\mathcal{Y}}$ represent Lagrange multipliers. $\mu$ and $\rho$ are the penalty parameters. We adopt the Augmented Lagrangian Multiplier (ALM) with the Alternative Direction Minimizing (ADM) optimization algorithm for solving the above optimization problem, and the updating rules of varying variables are as follows.

$\bullet$ \textbf{Updating $\bm{F}_S$ and $\bm{F}_A$.} For the convenient of optimization, we employ linear kernel $\bm{F}_S \bm{F}_A^T$ to construct the consensus graph $\bm{G}_{\bm{F}}$. By fixing the other variables, $\bm{F}_S$ and $\bm{F}_A$ can be updated by solving the following problem:
\begin{equation}
\label{eq:solve_F}
\begin{split}
&\mathop {\max}\limits_{ \bm{F}_S, \bm{F}_A } \sum_{v=1}^M { {(\bm{\alpha}^{(v)})}^{\gamma} tr\left(\bm{G}_{\bm{F}} \bm{G}^{(v)} {\bm{D}^{(v)}}^{-1}\right)}, \\
&s.t. \quad \bm{F}_S^T\bm{F}_S +  \bm{F}_A^T\bm{F}_A= \bm{I}.\\
\end{split}
\end{equation}
The Eq. (\ref{eq:solve_F}) has the closed-form solutions $\bm{F}_S=\frac{\sqrt{2}}{2}\bm{U}$ and $\bm{F}_A=\frac{\sqrt{2}}{2}\bm{V}$, in which $\bm{U}$ and $\bm{V}$ are the leading $d$ left and right singular vectors of the matrix $\sum_{v=1}^M {(\bm{\alpha}^{(v)})}^{\gamma}\bm{G}^{(v)} {\bm{D}^{(v)}}^{-1}$.

$\bullet$ \textbf{Updating $\bm{\mathcal{Z}}$.} In this case, $\bm{\mathcal{Z}}$ can updated by solving the following problem:
\begin{equation}\label{eq:solve_Z}
\begin{split}
&\mathop {\min}_{\bm{\mathcal{Z}}}  \frac{1}{2}\left\| \bm{\mathcal{G}} + \frac{1}{\rho} \bm{\mathcal{Y}}- \bm{\mathcal{Z}} \right\|_F^2 + \frac{\lambda_R}{\rho}\bm{\mathcal{Z}}_{\bm\omega, \ast}.\\
\end{split}
\end{equation}
The optimal solution of the Eq. (\ref{eq:solve_Z}) is $\bm{\Gamma}_{\frac{\lambda_R}{\rho}}[\bm{\mathcal{G}} + \frac{1}{\rho} \bm{\mathcal{Y}}]$. More details are placed in the \textbf{Appendix A}.

$\bullet$ \textbf{Updating $\bm{G}^{(v)}$.} In this case, $\bm{G}^{(v)}$ can updated by solving the following problem:
\begin{equation}\label{eq:solve_G}
\begin{split}
&\mathop {\min}_{\bm{G}^{(v)}}
    	\left \langle\bm{Y}^{(v)}, \bm{S}^{(v)}-\bm{G}^{(v)}-\bm{E}^{(v)}	\right \rangle + \frac{\mu}{2}\left\| \bm{S}^{(v)}-\bm{G}^{(v)}-\bm{E}^{(v)} \right\|_F^2    \\ & \qquad+ \left \langle \bm{\mathcal{Y}}^{(v)}, \bm{{G}}^{(v)} - \bm{\mathcal{Z}}^{(v)} \right \rangle + \frac{\rho}{2}\left\| \bm{{G}}^{(v)} - \bm{\mathcal{Z}}^{(v)} \right\|_F^2 \\ & \qquad
{ -{(\bm{\alpha}^{(v)})}^{\gamma} tr(\bm{G}_{\bm{F}} \bm{G}^{(v)} {\bm{D}^{(v)}}^{-1})}, \\
& s.t. \quad \bm{G}^{(v)} \ge 0, \quad \bm{G}^{{(v)}^T}\bm{1}=\bm{1},\\
\end{split}
\end{equation}
where $\bm{\mathcal{Z}}^{(v)}=\bm{\mathcal{Z}}(:,v,:)$ and $\bm{\mathcal{Y}}^{(v)}=\bm{\mathcal{Y}}(:,v,:)$. It can be shown that the above formula has a closed connection, and the corresponding proof is placed in the \textbf{Appendix B}.

$\bullet$ \textbf{Updating $\bm{E}^{(v)}$.} In this case, $\bm{E}^{(v)}$ can updated by solving the following problem:
\begin{equation}\label{eq:solve_E}
\begin{split}
    & \mathop {\min}\limits_{ \bm{E}^{(v)} }  \frac{1}{2}\left\| \bm{E}^{(v)} - \bm{\Gamma}^{(v)}\right\|_F^2 +  \frac{\lambda_E}{\mu} {\left\|\bm{E}^{(v)}\right\|_1},\\
\end{split}
\end{equation}
where $\bm{\Gamma^{(v)}}=\bm{S}^{(v)}-\bm{G}^{(v)}-\frac{1}{\mu}\bm{Y}^{(v)}$. Based on the proximal gradient-decent method, the optimal solution $\bm{E}^{(v)}$ of Eq. (\ref{eq:solve_E}) is $max(|\bm{\Gamma^{(v)}}|-\frac{\lambda_E}{\mu}, 0)$.

$\bullet$ \textbf{Updating $\bm\alpha^{(v)}$.} In this case, $\bm\alpha^{(v)}$ can be updated by solving the following problem:
\begin{equation}
\label{eq:solve_alpha}
\begin{split}
&\mathop {\max}\limits_{ \bm{\alpha} } \sum_{v=1}^M { {(\bm{\alpha}^{(v)})}^{\gamma} tr\left(\bm{G}_{\bm{F}} \bm{G}^{(v)} {\bm{D}^{(v)}}^{-1}\right)}, \\
&s.t. \quad \sum_{v=1}^M \bm{\alpha}^{(v)}=1,  \bm{\alpha}^{(v)} \ge 0.\\
\end{split}
\end{equation}
Using the Lagrange multiplier method, we can obtain the closed-form solutions of the Eq. (\ref{eq:solve_alpha}).  More details are placed in the \textbf{Appendix C}.

$\bullet$ \textbf{Updating Lagrange multipliers and penalty parameters.} Lagrange multipliers and penalty parameters can be updated as follows:
\begin{equation}
    \begin{split}
        & \bm{Y}^{(v)} := \bm{Y}^{(v)}+\mu (\bm{S}^{(v)}-\bm{G}^{(v)}-\bm{E}^{(v)}),\\
        & \bm{\mathcal{Y}} := \bm{\mathcal{Y}} + \rho (\bm{\mathcal{G}}-\bm{\mathcal{Z}}), \\
        & \mu := \eta \mu, \\
        & \rho := \eta \rho, \\
    \end{split}
\end{equation}
where $\eta > 1$ is used to boost the convergence speed \cite{chen2020graph}.

\subsubsection{Time Complexity Analysis}
In this part, the computational complexity analysis of solving the problem in Eq. (\ref{eq:lang}) is provided. The main computation consists of five parts, which are corresponding to the updating process in the optimization section. The time complexities in iteratively updating these variables are $\bm{O}(MNK+K^2N)$, $\bm{O}(MNKlog(MN)+M^2NK)$, $\bm{O}(MNKd+MNKlog(K))$, $\bm{O}(MNK)$, and $\bm{O}(MNK)$, respectively. Therefore, the main time complexity in each iteration is $\bm{O}(N(MK+K^2+MKlog(MN)+M^2K+MKd+MKlog(K)+2MK))$. While $K \ll N$, the main complexity of TCGF is linear to $N$.

\subsubsection{Convergence analysis}
Since the model in Eq. (\ref{eq:lang}) is not a joint convex problem of all variables, it still remains an open problem to require a globally optimal solution. Fortunately, by means of the alternating optimization algorithm, the proposed model can be solved. Due to the convex property and optimal solution of each sub-problem, the optimization can be shown to converge. In what follows, we would introduce the convergence of each sub-problem.

For updating $\bm{F}_S$ and $\bm{F}_A$, sub-problem in the Eq. (\ref{eq:solve_F}) is equal to the following equation:
\begin{equation}
\begin{split}
& \mathop {\max}\limits_{ \bm{F}\bm{F}^T=\bm{I} } \sum_{v=1}^M { {(\bm{\alpha}^{(v)})}^{\gamma} tr\left( \bm{F}\bm{L}^{(v)} \bm{F}^T\right)}, \\ 
\end{split}
\end{equation}
where $\bm{F} = [\bm{F}_S; \bm{F}_A] \in \mathcal{R}^{(N+K) \times d}$ and $\bm{L}^{(v)}=[\bm{0} \ (\bm{G}^{(v)} \bm{D}^{{(v)}^{-1}}); {(\bm{G}^{(v)} \bm{D}^{{(v)}^{-1}})}^T \ \bm{0}] \in \mathcal{R}^{(N+K) \times (N+K)} $. Obviously, the Hessian matrix $ \sum_{v=1}^M { {(\bm{\alpha}^{(v)})}^{\gamma} \bm{L}^{(v)}}$ of the above equation is positive semi-definite. Thus, sub-problem in the Eq. (\ref{eq:solve_F}) is strictly convex.

For updating $\bm{\mathcal{Z}}$, $\bm{\mathcal{Z}}=\bm{\Gamma}_{\frac{\lambda_R}{\rho}}[\bm{\mathcal{G}} + \frac{1}{\rho} \bm{\mathcal{Y}}]$ is a closed-form solution, thus the sub-problem in the Eq. (\ref{eq:solve_Z}) is a convex function.

For updating $\bm{G}^{(v)}$, the second order derivative of this function in Eq. (\ref{eq:solve_G}) with respect to $\bm{G}^{(v)}(i, :)$ is equal to 1, thus it's easy to check that the objective function of sub-problem in the eq. (\ref{eq:solve_G}) is also a convex function.

For updating $\bm{E}^{(v)}$, it's readily showed that the objective value in the Eq. (\ref{eq:solve_E}) is monotonically decreased due to the convergence property of the proximal gradient-decent method \cite{beck2009fast}.

For updating $\bm{\alpha}^{(v)}$, the sub-problem in the Eq. (\ref{eq:solve_alpha}) is a linear convex function, and a closed-form solution can be assigned to $\bm{\alpha}^{(v)}$.


\section{Experiments and Analysis}

In this section, we report the experimental results that have been conducted to evaluate the performance of the proposed TCGF model using seven real-world datasets. Additionally, we provide detailed analysis to illustrate the effectiveness and robustness of the proposed TCGF.

\begin{figure*}[htbp]
\centering
\subfigure[Caltech101 dataset.]{
\centering
\includegraphics[width=0.41\textwidth]{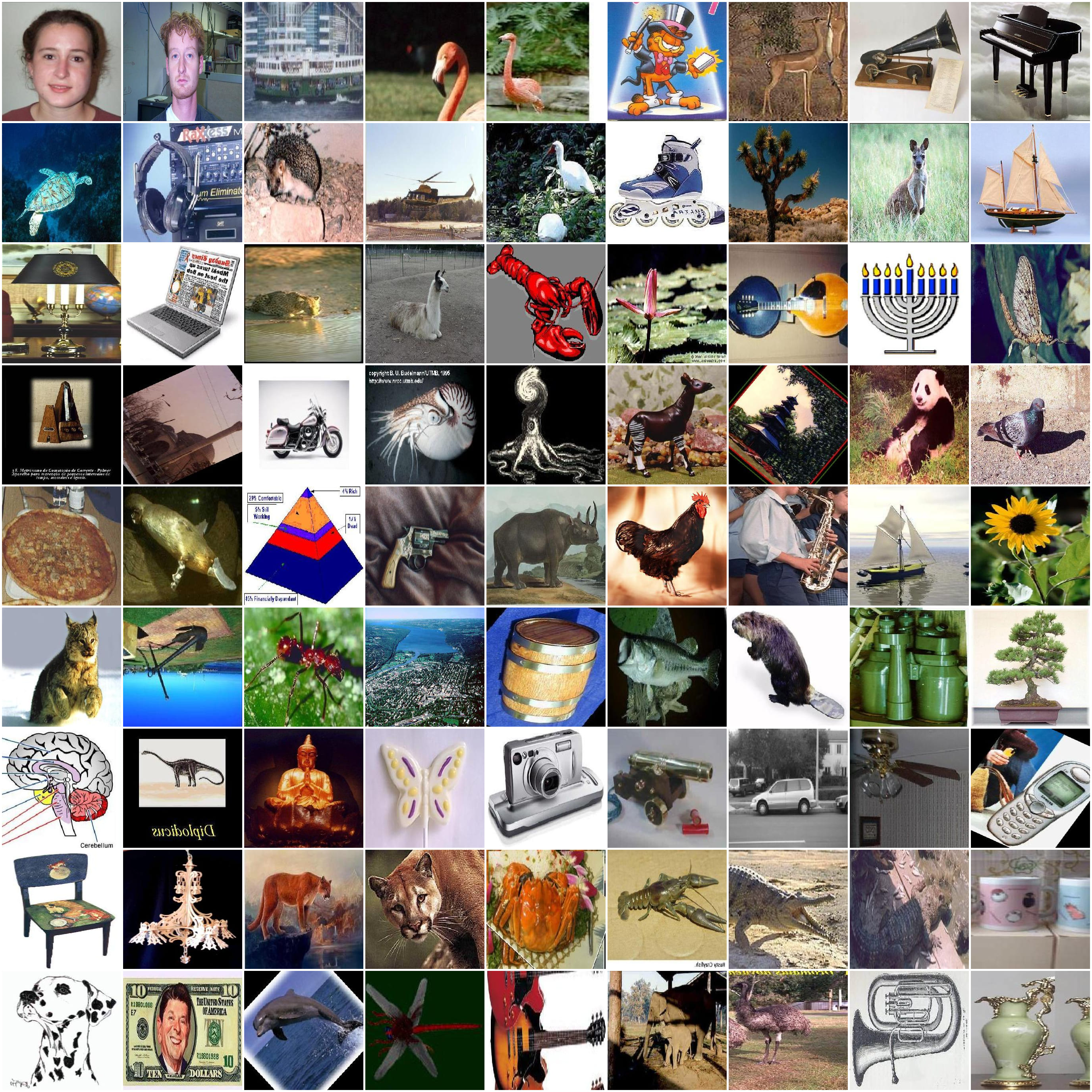}
}
\subfigure[ALOI dataset.]{
\centering
\includegraphics[width=0.55\textwidth]{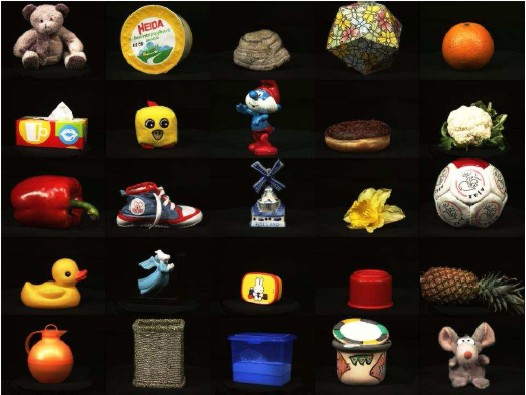}
}

\subfigure[MSRC dataset.]{
\centering
\includegraphics[width=0.43\textwidth]{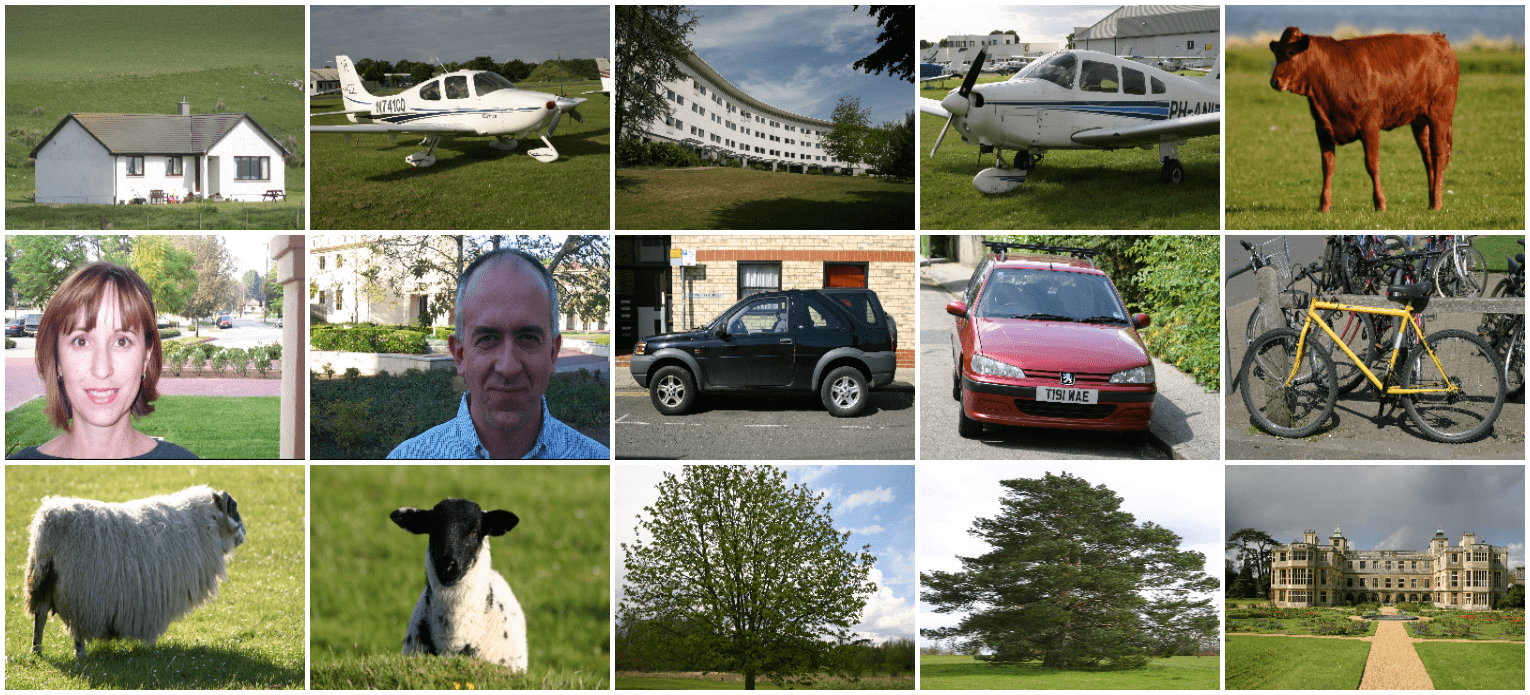}
}
\subfigure[YoutubeFace dataset.]{
\centering
\includegraphics[width=0.52\textwidth]{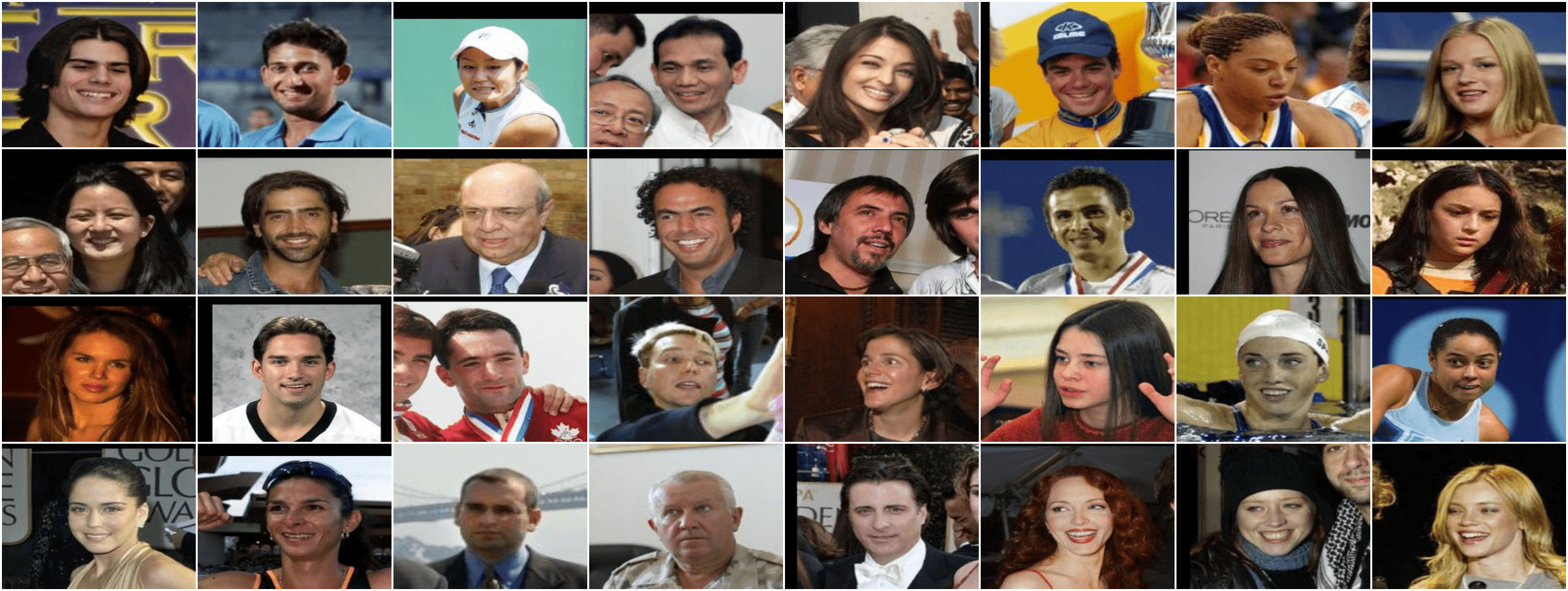}
}

\caption{Some examples images in datasets.}
\label{example_images}
\end{figure*}

\subsection{Experiment Settings}
\subsubsection{Datasets}
We evaluate our proposed framework on six benchmark datasets MSRC\footnote{http://archive.ics.uci.edu/ml}, NGs\footnote{http://lig-membres.imag.fr/grimal/data.html.}, Hdigit\footnote{https://cs.nyu.edu/roweis/data.html}, Caltech101\footnote{https://data.caltech.edu/records/mzrjq-6wc02/files/caltech-101.zip}, ALOI\_100\footnote{https://elki-project.github.io/datasets/multi\_view}, ALOI\_1K\footnote{https://elki-project.github.io/datasets/multi\_view} and YoutubeFace\footnote{https://www.cs.tau.ac.il/wolf/ytfaces/}. Ngs is a document dataset, and the resting datasets are image datasets, in which some samples in image datasets are shown in Fig. \ref{example_images}. The detailed information of these datasets is summarized as follows:
\begin{itemize}
    \item[1.] \textbf{MSRC} contains 240 samples from 8 classes, where each category includes 30 images. Following the work \cite{cai2011heterogeneous}, 7 classes containing Tree, building, airplane, cow, face, car and bicycle, are selected. Five image descriptors \cite{wu2010centrist} are utilized to extract multi-view features, which are colour moment with dimension 24, local binary pattern with dimension 256, HOG with dimension 576, GIST with dimension 512 and CENTRIST with dimension 254.

    \item[2.] \textbf{NGs} is the subset of the 20 Newsgroups data set that is a collection of approximately 20,000 newsgroup documents, partitioned (nearly) evenly across 20 different newsgroups. It consists of 500 newsgroup documents. Each raw document was pre-processed with three different methods (giving three views), and was annotated with one of five topical labels.

    \item[3.] \textbf{Hdigit} contains 10000 handwritten digit images from two sources, including MNIST and USPS Handwritten Digits. These digit images belong to 10 subjects (0-9). In our experiments, MNIST and USPS Handwritten Digits can be used as two views, whose dimensions are 784 and 256, respectively.

    \item[4.] \textbf{Caltech101} contains 9144 images of 101 categories. Following related multi-view clustering works, we select 20 categories with 2,386 images as dataset. Six image descriptors can be employed to extract multi-view features for each image, including GABOR with dimension 48, WM with dimension 48, CENT with dimension 254, HOG with dimension 984, GIST with dimension 512 and local binary pattern with dimension 256.

    \item[5.] \textbf{ALOI\_1K} contains 110250 images of 1000 small objects. For each image, we use three image descriptors to extract its features as 3 views, which are RGB color histograms with dimension 77, HSV/HSB color histograms with dimension 13, color affinity with dimension 64, Haralick feature with dimension 125.

    \item[6.] \textbf{ALOI\_100} is a subset of ALOI\_1K, in which We select 100 classes with three views. For each category, 1080 images are selected, thus it contains 10800 images in our experiments.

    \item[7.] \textbf{YoutubeFace} contains complicated backgrounds, dynamic postures, and various obstacles, which is produced from YouTube with 101499 face images of 31 groups. Five image descriptors are utilized to extract their features as multi-view features.

\end{itemize}
For the convenient of readers, the statistical characteristics of six used multi-view datasets can be summarized in Table \ref{tab:datasets}.

\begin{table}[htbp]
\caption{The statistic information of eight real-world datasets.}
\label{tab:datasets}
\centering
\begin{tabular*}{0.48\textwidth}{@{\extracolsep{\fill}}lcccc}
\hline
Dataset & Samples & Classes & Views & Dimensions\\
\hline
MSRC & 210 & 7 & 5 & 24$\backslash$256$\backslash$576$\backslash$512$\backslash$254 \\
NGs & 500 & 20 & 3 & 2000$\backslash$2000$\backslash$2000 \\
Caltech101 & 2386 & 6 & 20 & 48$\backslash$40$\backslash$254$\backslash$1984$\backslash$512$\backslash$928\\
Hdigit & 10000 & 10 & 2 & 784$\backslash$256 \\
ALOI\_100 & 10800 & 100 & 4 & 77$\backslash$13$\backslash$64$\backslash$125 \\
ALOI\_1K & 110250 & 1000 & 4 & 77$\backslash$13$\backslash$64$\backslash$125 \\
YoutubeFace & 101499 & 31 & 5 & 64$\backslash$512$\backslash$64$\backslash$647$\backslash$838\\
\hline
\end{tabular*}
\end{table}

\begin{table*}[htbp]
\small
\caption{Clustering performance comparisons in terms of ACC on seven real-world datasets.}
\label{tab:acc}
\centering
\renewcommand\arraystretch{1.5}
\begin{tabular*}{0.98\textwidth}{@{\extracolsep{\fill}}lccccccccccccc}  
\hline
Datasets & SC & GNMF & Co-reg & MDcR & AMGL & GMC & GFSC & LMVSC & FPMVS & VCGA & LTBPL & TCGF\\
\hline
\multicolumn{13}{c}{Small-scale multi-view datasets} \\
\hline
MSRC & 0.776 & 0.148 & 0.814 & 0.468 & 0.738 & 0.748 & 0.742 & 0.343 & 0.786 & \textbf{0.834} & 0.705 & \textbf{1.000} \\
NGs & 0.631 & 0.296 & 0.368 & 0.590 & 0.308 & 0.982 & 0.511 & 0.298 & 0.738 & 0.980 & \textbf{0.994} & \textbf{1.000} \\
\hline
\multicolumn{13}{c}{General-scale multi-view datasets} \\
\hline
Hdigit & 0.526 & 0.935 & 0.816 & 0.866 & 0.901 & \textbf{0.998} & 0.895 & 0.854 & 0.910 & 0.918 & \textbf{0.999} & \textbf{1.000} \\
Caltech101 & 0.442 & 0.439 & 0.412 & 0.465 & 0.557 & 0.456 & 0.702 & 0.384 & 0.655 & - & 0.694 & \textbf{0.789} \\
ALOI\_100 & 0.492 & 0.261 & 0.521 & 0.279 & 0.600 & 0.571 & 0.281 & 0.566 & 0.342 & - & \textbf{0.626} & \textbf{0.673} \\
\hline
\multicolumn{13}{c}{\underline{Large-scale multi-view datasets}} \\
ALOI\_1K & 0.209 & 0.152 & OOM & OOM & OOM & OOM & OOM & \textbf{0.366} & 0.218 & OOM & OOM & \textbf{0.411} \\
YoutubeFace & 0.139 & 0.092 & OOM & OOM & OOM & OOM & OOM & 0.148 & \textbf{0.241} & OOM & OOM & \textbf{0.267} \\
\hline

\end{tabular*}
\end{table*}

 \subsubsection{Compared Methods}
In order to show the excellent performance of the proposed TCGF, this paper introduces nine state-of-the-art multi-view algorithms as comparing methods, including \textbf{Co-reg} \cite{sharma2012generalized}, \textbf{MDcR} \cite{Zhang2016Flexible}, \textbf{AMGL} \cite{Nie2017Auto}, \textbf{GMC} \cite{wang2019gmc}, \textbf{GFSC} \cite{kang2020multi}, \textbf{LMVSC} \cite{kang2020large}, \textbf{FPMVS} \cite{wang2021fast}, \textbf{VCGA} \cite{gu2022individuality} and \textbf{LTBPL} \cite{chen2022low}. Moreover, two typical single-view methods \textbf{SC} \cite{Ulrike2007Spectral} and \textbf{GNMF} \cite{Cai2011Graph} are adopted to show the advantages of multi-view clustering algorithms, which utilize the most informative view. The details of these compared methods can be summarized as follows:
\begin{itemize}
    \item[1.] \textbf{SC} is a standard spectral clustering method for clustering each single view, which is used for recognizing the different confidences of different views.

    \item[2.] \textbf{GNMF} is a method that attempts to explore a matrix factorization to respect the graph structure, and then consider the geometric structure in the data.

    \item[3.] \textbf{Co-reg} is a multiview spectral clustering method proposed in work, which regularizes different views to be close to each other.

    \item[4.] \textbf{MDcR} is a multi-view dimensionality reduction method, which explores the correlations of different views based on HSIC term.

    \item[5.] \textbf{AMGL} is an auto-weighted multiple graph learning method, which could allocate ideal weight for each view automatically.

    \item[6.] \textbf{GMC} is a multi-view graph-based method to learn the common graph shared by all views, and  directly gives the final clusters.

    \item[7.] \textbf{GFSC} is a multi-view spectral embedding based on multi-graph fusion to approximate the original graph of individual view.

    \item[]8.] \textbf{LMVSC} first learn a smaller graph for each view, and then integrates those graphs to transform original multi-view problems into single-view scenario.

    \item[9.] \textbf{FPMVS} jointly learns anchor selection and subspace graph construction into a unified optimization formulation to promote clustering quality, which can automatically learn an optimal anchor subspace graph without any extra hyper-parameters.

    \item[10.] \textbf{VCGA} first constructs the view-specific graphs and the shared graph from original multi-view data and hidden latent representation, and then the view-specific graphs of different views and the consensus graph are aligned into an informative target graph.

    \item[11.] \textbf{LTBPL} stacks multiple low-rank probability affinity matrices in a low-rank constrained tensor to recover their comprehensiveness and higher-order correlations, and links consensus indicator graph with view-specific representation carrying different adaptive confidences.

\end{itemize}


For a fair comparison, we download the released code of comparison algorithms from their original websites. Since all methods need to utilize k-means or connected component algorithms to get the final clustering results, which can be disturbed by the initialized states. Thus, we repeatedly run 10 times clustering experiments to eliminate the randomness in initialization for all compared methods and report the average performance. For the selection for hyper-parameters, the details od the proposed TCGF are placed in the section \ref{sub:hyper-ana}, and we tune these hyper-parameters following by corresponding papers for compared methods.



\subsubsection{Evaluation Metrics}
Various metrics have been proposed from diverse perspectives to evaluate the quality of the obtained
data clusters. In general, larger metrics indicate better clustering performances. To facilitate a comprehensive comparison, three metrics \cite{xie2018hyper} - accuracy (ACC), normalized mutual information (NMI), and purity (PUR) - that are commonly used in the clustering field.





\subsection{Experimental Results and Analysis}
Performance comparison of different methods. The highest performance is highlighted in boldface. The best two scores are highlighted in bold. ’OOM’ means out of memory. From experimental results in Tables \ref{tab:acc}-\ref{tab:pur}, we have the following observations.
\begin{itemize}
    \item The proposed TCGF achieves the best performance in terms of three metrics against other counterparts in most circumstances. Taking the results on MSRC, NGs, and Hdigit datasets for instances, TCGF has been considered as the strongest multi-view learning algorithm, which obtained 100\% in terms of ACC, NMI and PUR. This indicates that TCGF can be more effective and suitable for multiview features, which can well uncover the intrinsic rich information in multi-view features.

    \item For the ALOI\_1K and YoutubeFace datasets that contains over 100,000 samples, most of multi-view learning methods suffer from out-of-memory errors, such as GMC and LTBPL. The main reason is that the computational complexity of these works is squared or cubic with the data size. However, the proposed TCGF is capable of handling such large-scale datasets, which can obtain comparable and better performance against LMVSC and FPMVS. Meanwhile, the proposed TCGF performs these two methods in other datasets. It implies the effectiveness and efficiency of TCGF, which is applicable on both normal- and large-scale multi-view datasets.

\end{itemize}


\begin{table*}[htbp]
\small
\caption{Clustering performance comparisons in terms of NMI on seven real-world datasets.}
\label{tab:nmi}
\centering
\renewcommand\arraystretch{1.5}
\begin{tabular*}{0.98\textwidth}{@{\extracolsep{\fill}}lccccccccccccc}  
\hline
Datasets & SC & GNMF & Co-reg & MDcR & AMGL & GMC & GFSC & LMVSC & FPMVS & VCGA & LTBPL & TCGF\\
\hline
\multicolumn{13}{c}{Small-scale multi-view datasets} \\
MSRC & 0.625 & 0.023 & 0.719 & 0.402 & 0.714 & \textbf{0.771} & 0.668 & 0.246 & 0.686 & 0.751 & 0.628 & \textbf{1.000} \\
NGs & 0.562 & 0.105 & 0.119 & 0.501 & 0.191 & 0.939 & 0.451 & 0.101 & 0.592 & 0.935 & \textbf{0.979} & \textbf{1.000} \\
\hline
\multicolumn{13}{c}{General-scale multi-view datasets} \\
Hdigit & 0.468 & 0.862 & 0.689 & 0.777 & 0.883 & \textbf{0.993} & 0.854 & 0.808 & 0.816 & 0.849 & \textbf{0.999} & \textbf{1.000} \\
Caltech101 & 0.269 & 0.378 & 0.587 & 0.564 & 0.385 & 0.552 & \textbf{0.815} & 0.283 & 0.632 & - & 0.577 & 0.806 \\
ALOI\_100 & 0.687 & 0.466 & 0.699 & 0.536 & 0.681 & 0.735 & 0.542 & \textbf{74.95} & 0.648 & - & 0.655 & \textbf{0.762} \\
\hline
\multicolumn{13}{c}{Large-scale multi-view datasets} \\
ALOI\_1K & 0.544 & 0.530 & OOM & OOM & OOM & OOM & OOM & \textbf{0.704} & 0.555 & OOM & OOM & \textbf{0.679} \\
YoutubeFace & 0.121 & 0.071 & OOM & OOM & OOM & OOM & OOM & \textbf{0.133} & \textbf{0.243} & OOM & OOM & 0.102 \\
\hline  %

\end{tabular*}
\end{table*}

\begin{table*}[htbp]
\small
\caption{Clustering performance comparisons in terms of PUR on seven real-world datasets.}
\label{tab:pur}
\centering
\renewcommand\arraystretch{1.5}
\begin{tabular*}{0.98\textwidth}{@{\extracolsep{\fill}}lccccccccccccc}  
\hline
Datasets & SC & GNMF & Co-reg & MDcR & AMGL & GMC & GFSC & LMVSC & FPMVS & VCGA & LTBPL & TCGF\\
\hline
\multicolumn{13}{c}{Small-scale multi-view datasets} \\
\hline
MSRC & 0.627 & 0.171 & 0.713 & 0.492 & 0.795 & 0.790 & 0.761 & 0.381 & 0.786 & \textbf{0.835} & 0.714 & \textbf{1.000} \\
NGs & 0.689 & 0.328 & 0.368 & 0.693 & 0.330 & 0.982 & 0.504 & 0.818 & 0.738 & 0.980 & \textbf{0.994} & \textbf{1.000} \\
\hline
\multicolumn{13}{c}{General-scale multi-view datasets} \\
\hline
Hdigit & 0.345 & 0.935 & 0.660 & 0.866 & 0.912 & \textbf{0.996} & 0.914 & 0.875 & 0.910 & 0.918 & \textbf{0.999} & \textbf{1.000} \\
Caltech101 & 0.492 & 0.607 & 0.754 & 0.515 & 0.677 & 0.555 & 0.631 & 0.435 & 0.736 & - & 0.725 & \textbf{0.877} \\
ALOI\_100 & 0.493 & 0.288 & 0.519 & 0.512 & 0.634 & 0.507 & \textbf{0.753} & 0.598 & 0.348 & - & 0.640 & \textbf{0.687} \\
\hline
\multicolumn{13}{c}{Large-scale multi-view datasets} \\
\hline
ALOI\_1K & 0.238 & 0.193 & OOM & OOM & OOM & OOM & OOM & \textbf{0.426} & 0.242 & OOM & OOM & \textbf{0.435} \\
YoutubeFace & 0.280 & 0.267 & OOM & OOM & OOM & OOM & OOM & 0.282 & \textbf{0.328} & OOM & OOM & \textbf{0.302} \\
\hline  %

\end{tabular*}
\end{table*}

\subsection{Ablation Study}
In this section, the ablation study is conducted to evaluate the effects of consensus graph learning, view-specific graph learning, and tersorized graph learning. Specifically, for the each test, the corresponding term is removed while retaining the other terms. For notation simplicity, we denote these three tests as TCGF-v1, TCGF-v2, and TCGF-v3, respectively. These tests are performed on MSRC and NGs datasets, and the results of clustering preformance comparison in terms of ACC, NMI and PUR are reported in Table \ref{tab:abldation}. According to the table, we can observe that TCGF achieves superior clustering performance compared with its variants in all testing cases. To this end, the ablation study demonstrates the necessity of the proposed model, which simultaneously takes view-specific graph learning, tersorized graph learning, and consensus graph learning into consideration.

\begin{table}[htbp]
\small
\caption{Comparison results of TCGF and its variants.}
\label{tab:abldation}
\centering
\renewcommand\arraystretch{1.5}
\begin{tabular*}{0.42\textwidth}{@{\extracolsep{\fill}}lcccc}  
\hline
Datasets & Variants & ACC &  NMI  & RUR \\
\hline
\multirow{4}*{MSCRv1} & TCGF-v1 & 0.910 & 0.821 & 0.910\\
& TCGF-v2 & 0.971 & 0.942 & 0.971\\
& TCGF-v3 & 0.819 & 0.728 & 0.819 \\
& TCGF & 1.000 & 1.000 & 1.000 \\
\hline
\multirow{4}*{NGs} & TCGF-v1 & 0.878 & 0.791 & 0.878\\
& TCGF-v2 & 0.952 & 0.866 & 0.952 \\
& TCGF-v3 & 0.664 & 0.592 & 0.696 \\
& TCGF & 1.000 & 1.000 & 1.000 \\
\hline
\end{tabular*}
\end{table}

\subsection{Hyper-parameter Analysis}
\label{sub:hyper-ana}
\begin{figure}[htbp]
\centering
\includegraphics[width=0.24\textwidth]{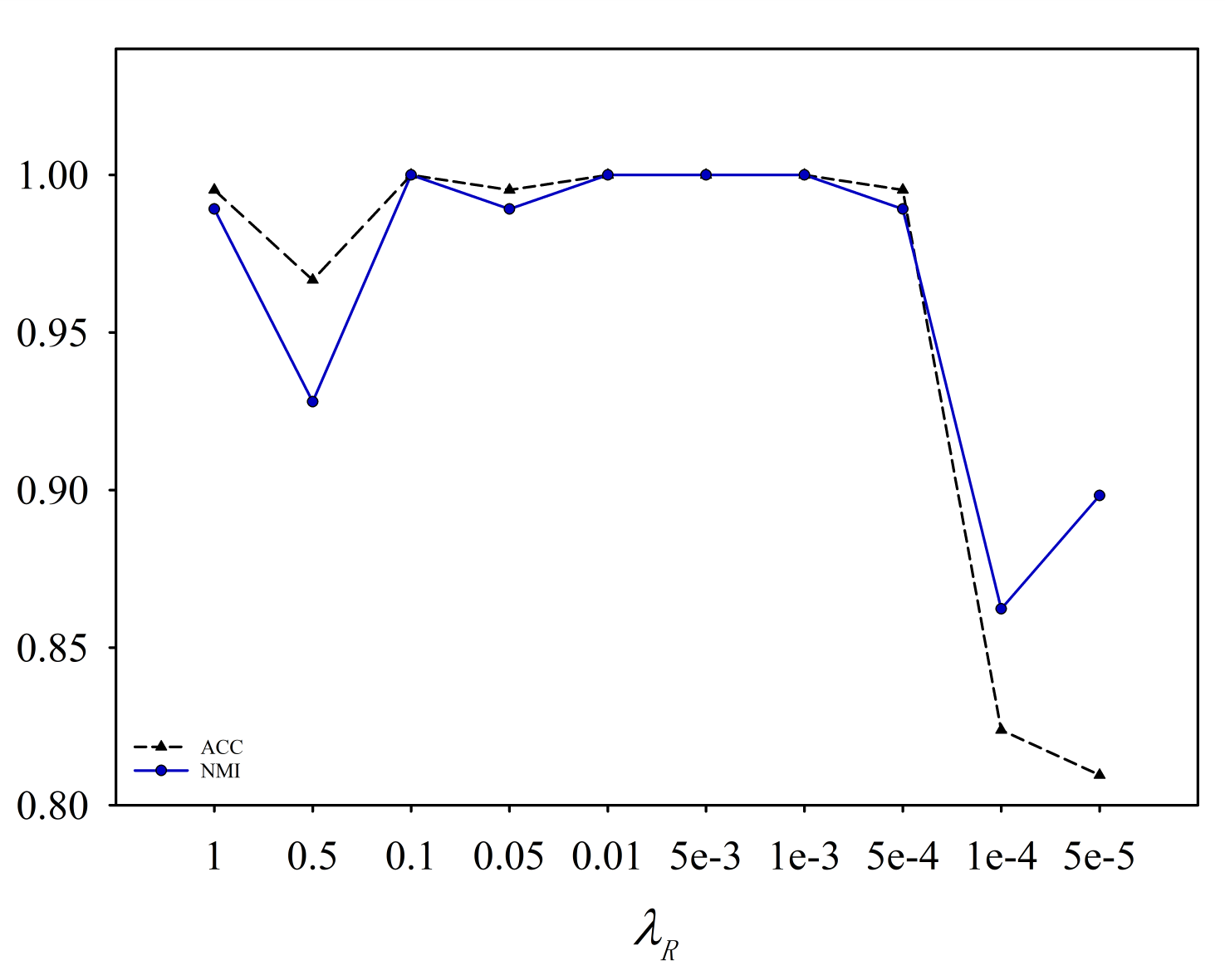}
\includegraphics[width=0.24\textwidth]{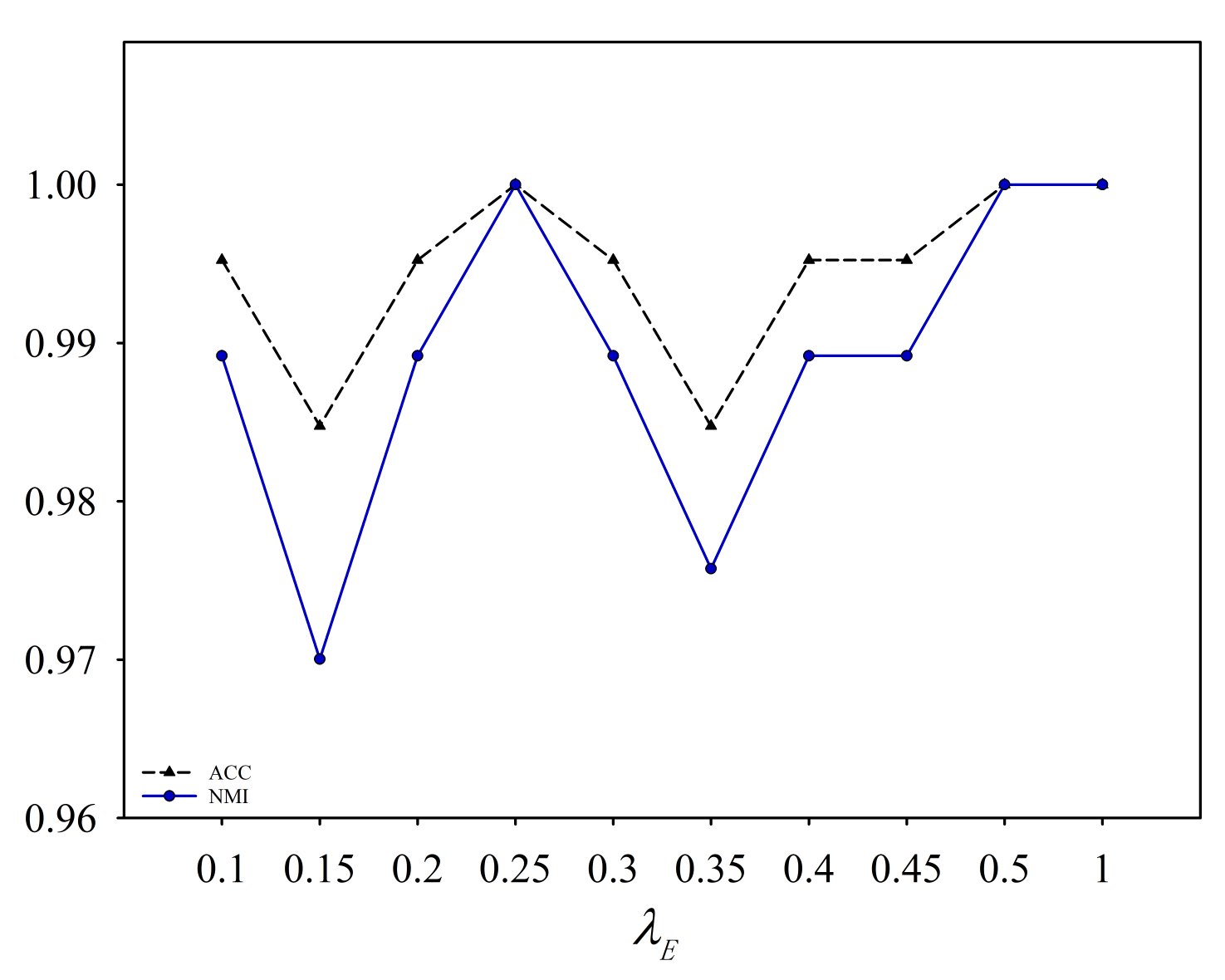}
\caption{Hyper-parameter analysis on MSRC dataset.}
\label{fig:MSRC-param}
\end{figure}

\begin{figure}[htbp]
\centering
\includegraphics[width=0.23\textwidth]{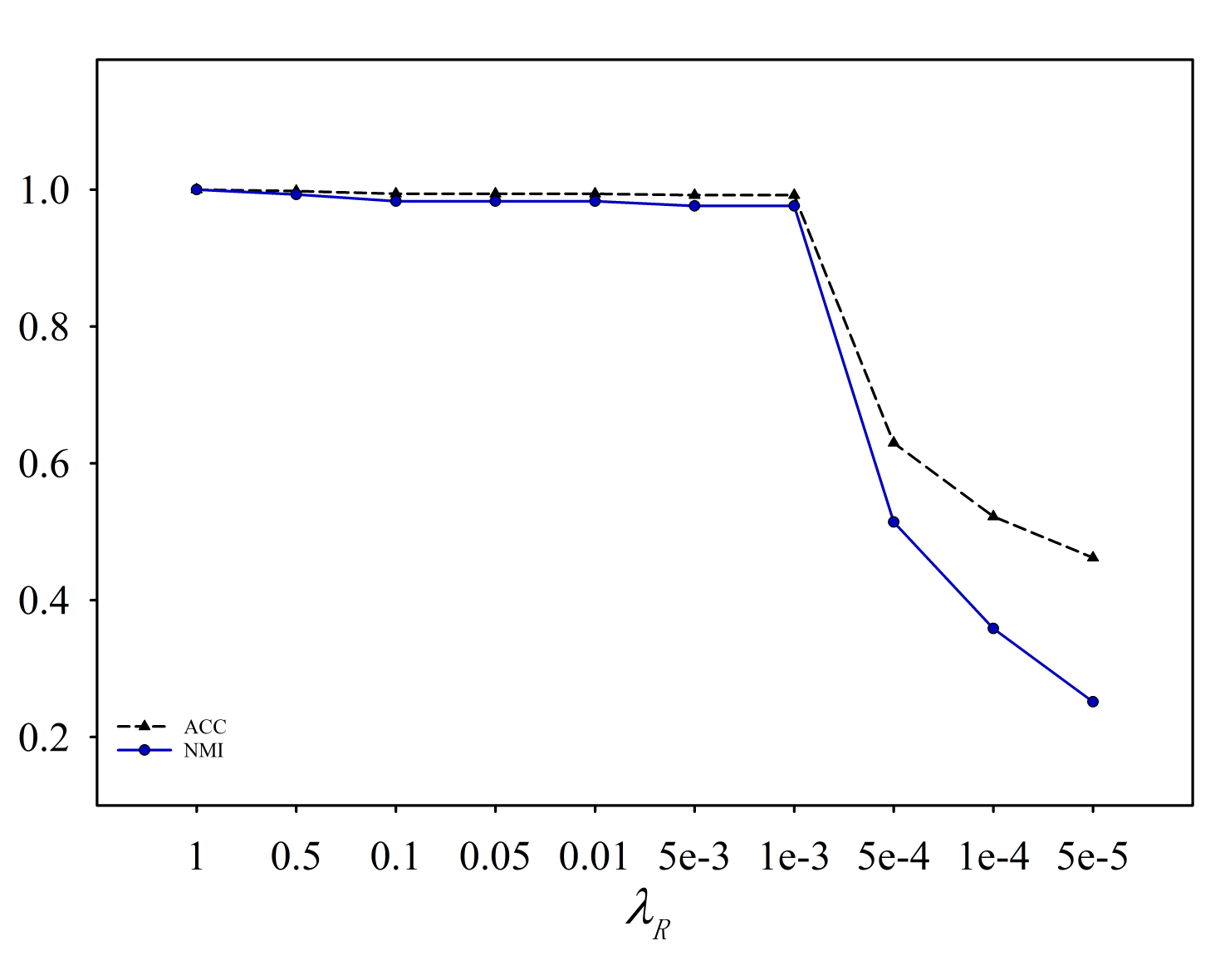}
\includegraphics[width=0.252\textwidth]{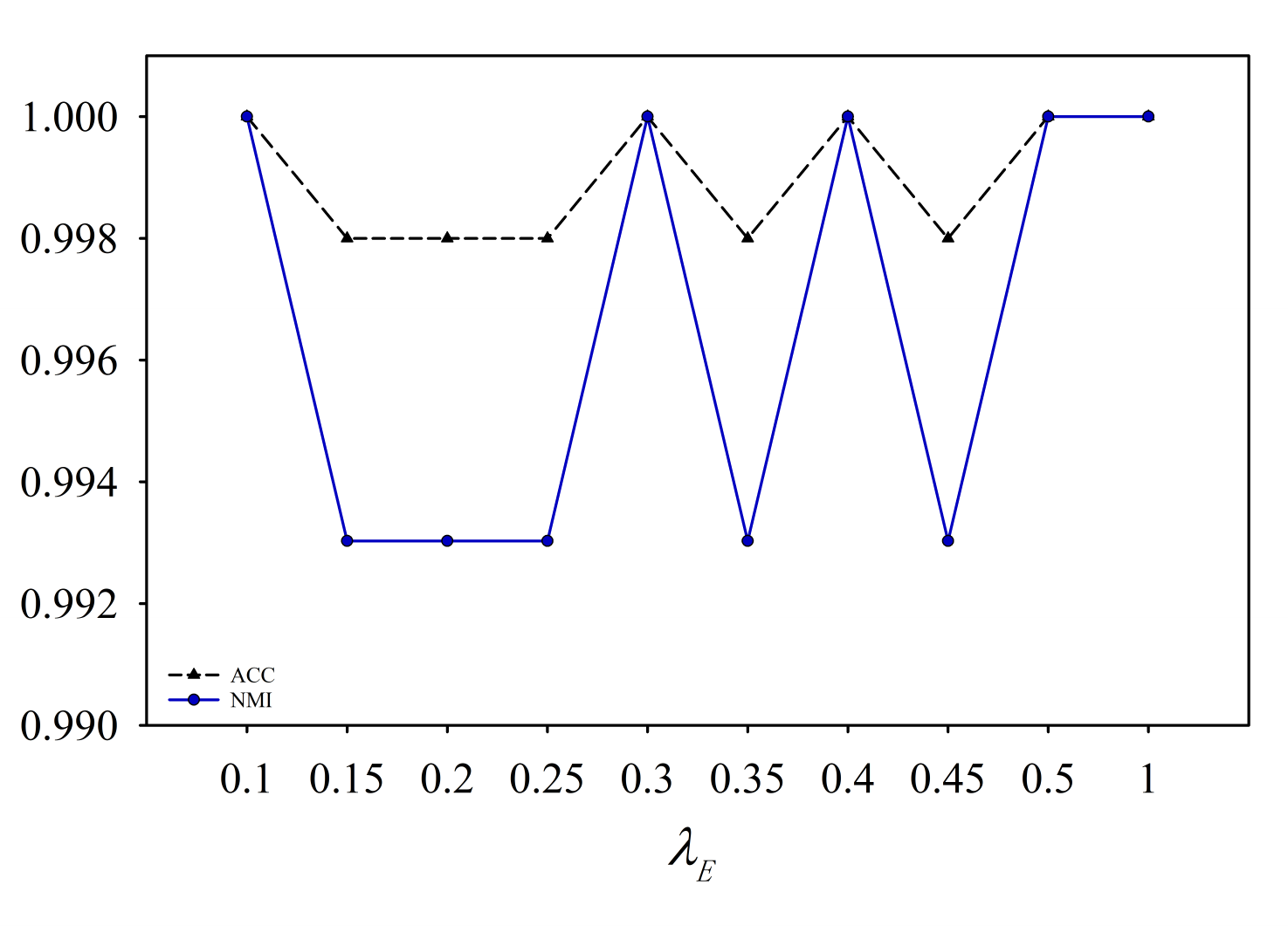}
\caption{Hyper-parameter analysis on NGs dataset.}
\label{fig:NGs-param}
\end{figure}

In this subsection, hyper-parameter analysis is conducted to investigate the effects of two parameters $\lambda_E$ and $\lambda_R$ on MSRC and NGs datasets with different settings, in which the experimental results in terms of ACC and NMI are reported in Figs. \ref{fig:MSRC-param}-\ref{fig:NGs-param}. Through these experimental results in Figs. \ref{fig:MSRC-param}-\ref{fig:NGs-param}, we can observe that for different datasets, the selections
of $\lambda_E$ and $\lambda_R$ are different, and the best values of
these two parameters vary from one dataset to another. However, there exists a wide range for each hyper-parameter in which relatively stable and good results can be readily obtained. Meanwhile, We set $\lambda_E$ and $\lambda_R$ to those values that make the proposed TCGF has the best results according to the experiments.

\subsection{Visualization}
\label{sub:vis}
\begin{figure*}[htbp]
\centering

\subfigure[Original Features in MSCRv1]{
\centering
\includegraphics[width=0.29\textwidth]{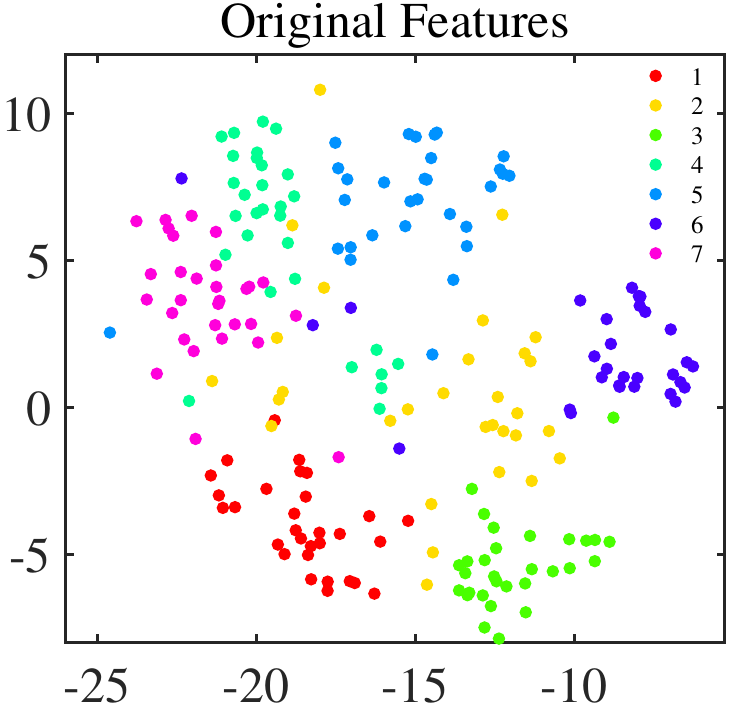}
}
\subfigure[Original Features in NGs]{
\centering
\includegraphics[width=0.31\textwidth]{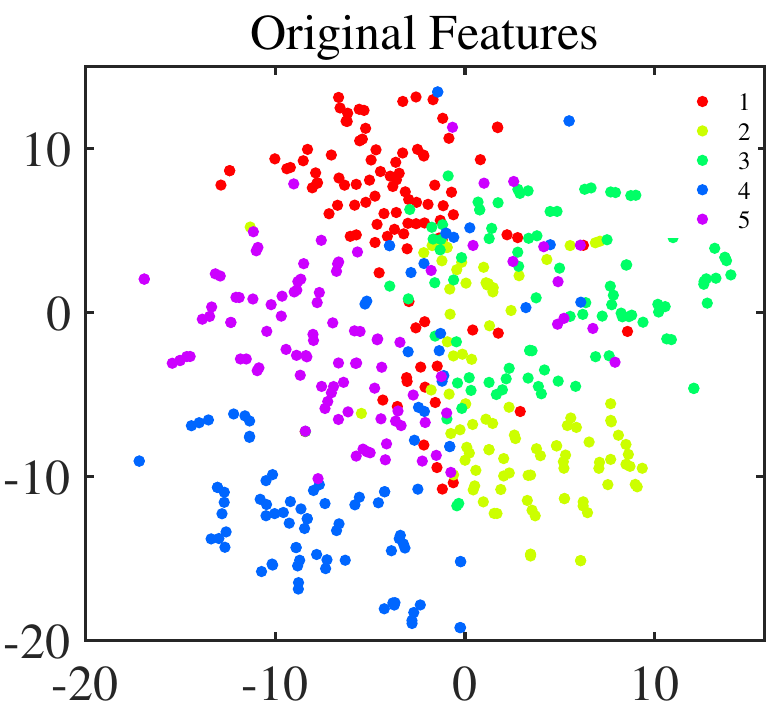}
}
\subfigure[Original Features in Hdigits]{
\centering
\includegraphics[width=0.335\textwidth]{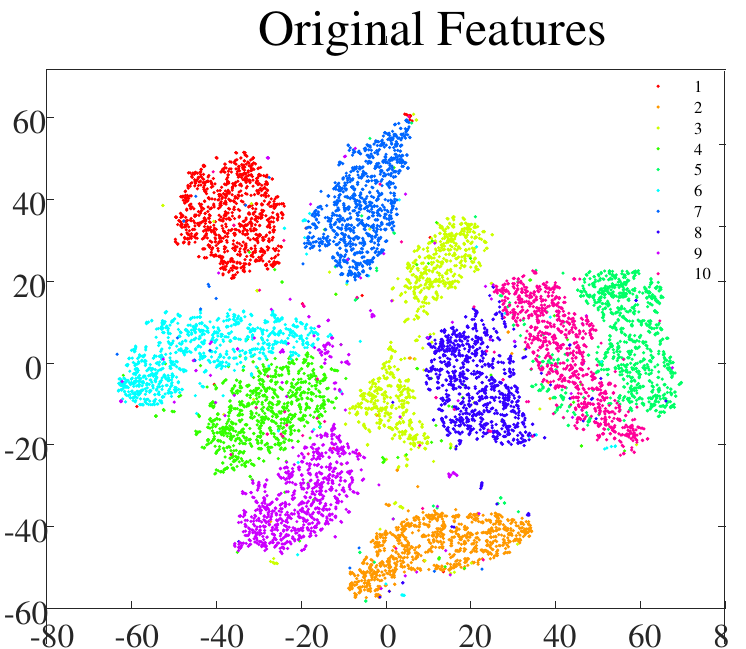}
}

\centering
\subfigure[Learnt Embedding by TCGF in MSCRv1]{
\centering
\includegraphics[width=0.295\textwidth]{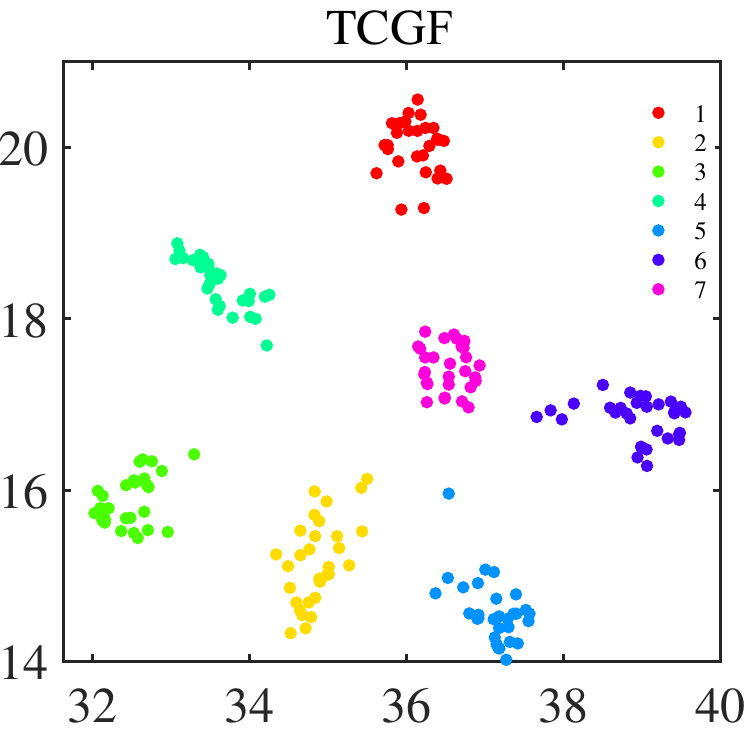}
}
\subfigure[Learnt Embedding by TCGF in NGs]{
\centering
\includegraphics[width=0.31\textwidth]{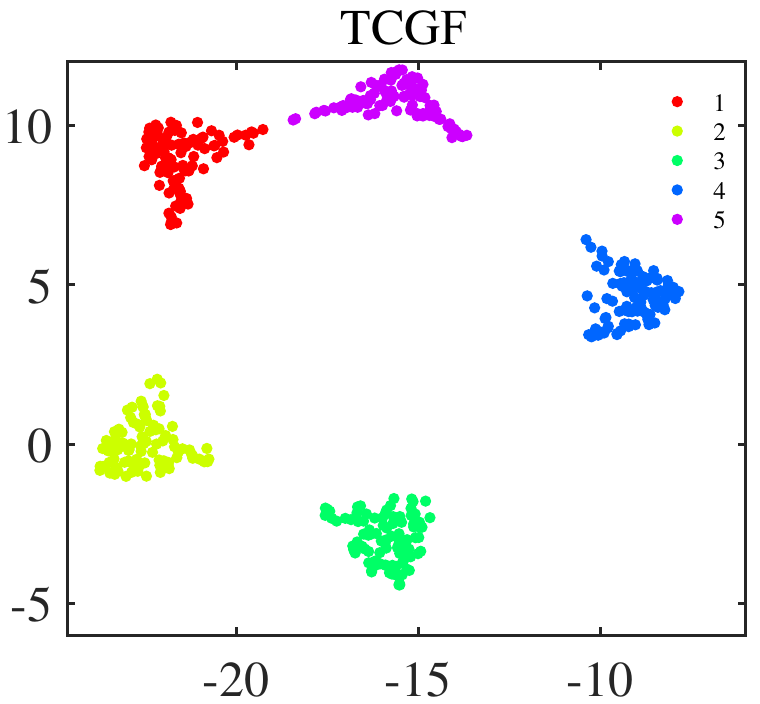}
}
\subfigure[Learnt Embedding by TCGF in Hdigits]{
\centering
\includegraphics[width=0.335\textwidth]{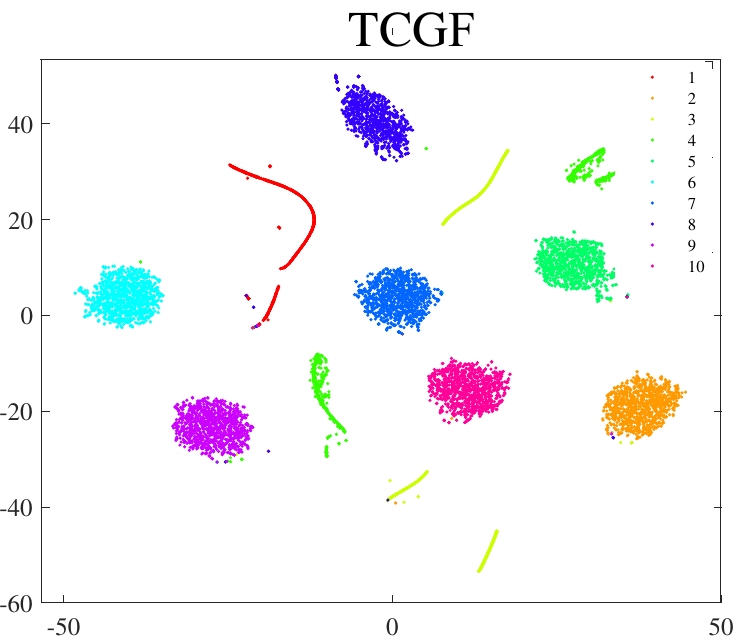}
}

\caption{Visualization of Original Features and Learnt Embedding on MSCRv1, NGs, and Hdigits datasets.}
\label{fig:NGs-vis}
\end{figure*}

Additional visualization results conducted on MSRC, NGs and Hdigits datasets are shown in Fig.\ref{fig:NGs-vis}, which adopt t-SNE \cite{van2008visualizing} to project features into the 2-dimensional subspace. Obviously, the distributions of original data are disordered. After TCGF is conducted, samples can be readily separated into several clusters, which further validates the effectiveness of TCGF.

\begin{figure*}[htbp]
\centering
\subfigure[MSCRv1]{
\centering
\includegraphics[width=0.3\textwidth]{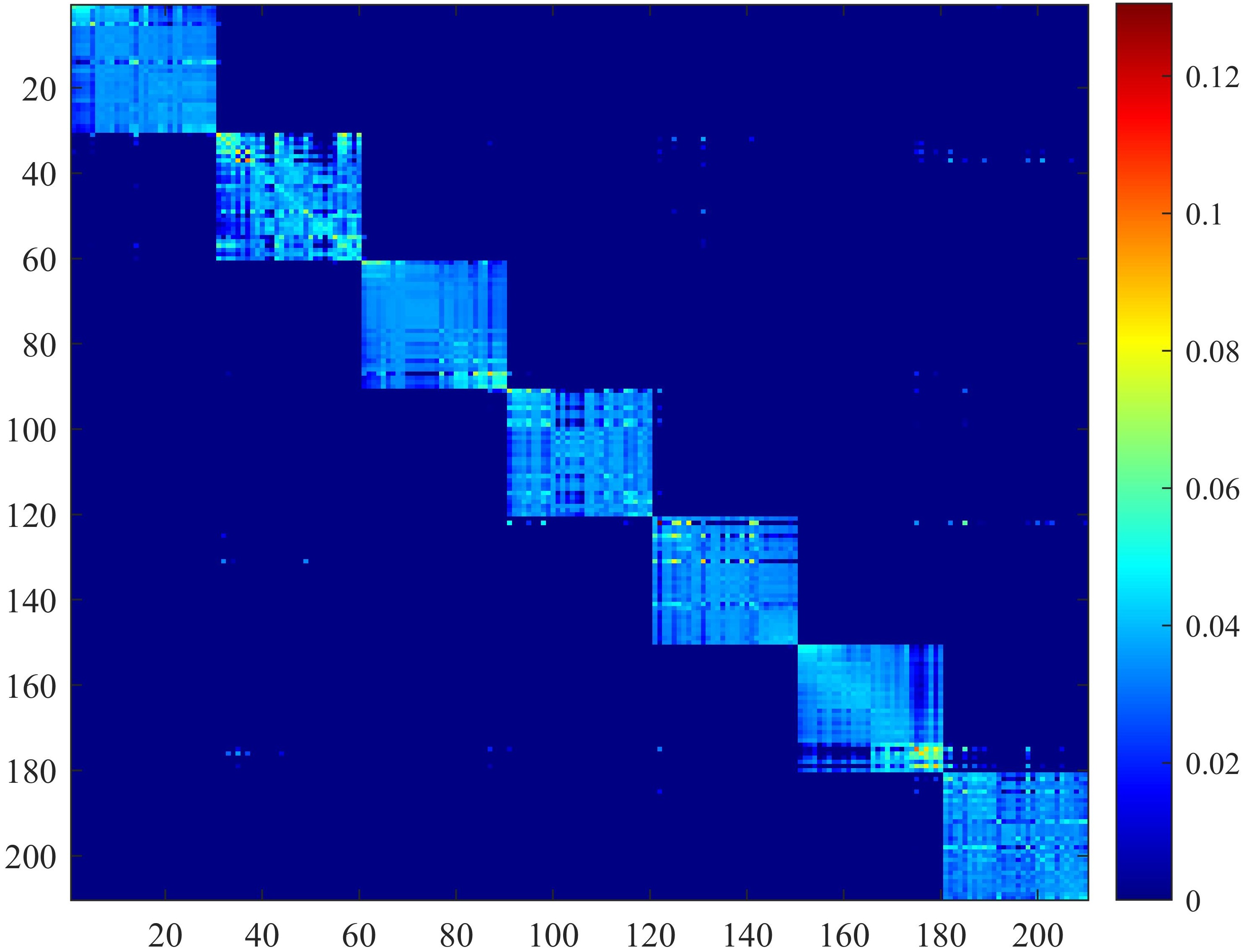}
}
\subfigure[NGs]{
\centering
\includegraphics[width=0.3\textwidth]{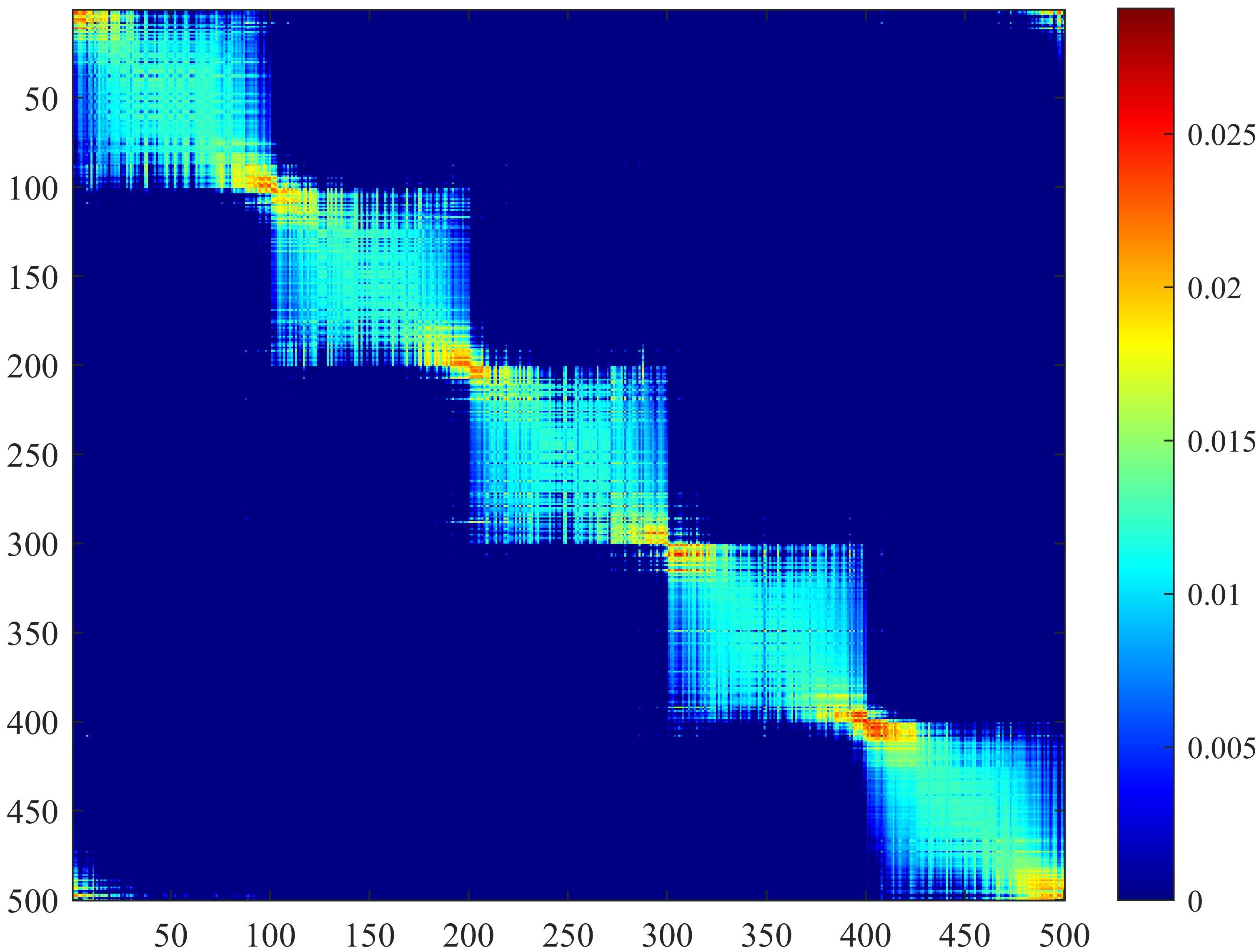}
}
\subfigure[Hdigits]{
\centering
\includegraphics[width=0.3\textwidth]{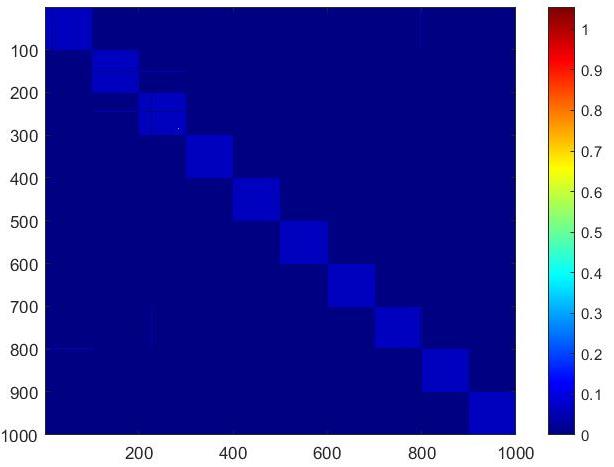}
}
\caption{The Learned Consensus Graphs of TCGF on MSCRv1, NGs, and Hdigits Datasets.}
\label{fig:graph-vis}
\end{figure*}

To show the superiority of the learned consensus graph, we visualized the consensus graphs of TCEF on MSRC, NGs and Hdigits datasets, as shown in Fig. \ref{fig:graph-vis}. As shown in Fig. \ref{fig:graph-vis},  we can find clear and complete block diagonal structures. Thus, TCEF can adaptively promote the learning of consensus graphs towards better attributes. Although there are some ineluctable noisy values in the consensus graphs, TCEF can achieve good results on multi-view clustering due to the superior attributes of the learned consensus graph. In this way, we can verify that how to construct a consensus graph is of vital importance for multi-view clustering.

\section{Conclusion}\label{conclusion}
In this paper, we propose a novel unified consensus embedding framework for multi-view representation learning, termed Tensorized Consensus Graph Framework (TCGF). TCGF aims to be served as a universal learning framework for existing multi-view works under arbitrary assumptions and is applicable for multi-view applications with varying scales. TCGF firstly proprovides a unified framework to exploit the representations for individual view, enabling it to be suitable for different-scales datasets. Then, it stacks these representations into a tensor under alignment basics as a high-order representation, allowing for the smooth propagation of consistency and complementary information across all views. Additionally, TCGF proposes learning a consensus embedding shared by adaptively collaborating all views to uncover the essential structure of the multi-view data, which utilizes view-consensus grouping effect to regularize the view-consensus representation. The proposed TCGF is evaluated through extensive experiments on seven different-scales datasets, demonstrating its effectiveness and superiority to maintain or outperform other sota multi-view methods.

\section*{Acknowledgements}
The authors would like to thank the anonymous reviewers for their insightful comments and suggestions to significantly improve the quality of this paper.

\bibliographystyle{IEEEtran}
\bibliography{IEEEexample}

\vspace{12pt}

\begin{IEEEbiography}[{\includegraphics[width=1in,height=1.25in,clip,keepaspectratio]{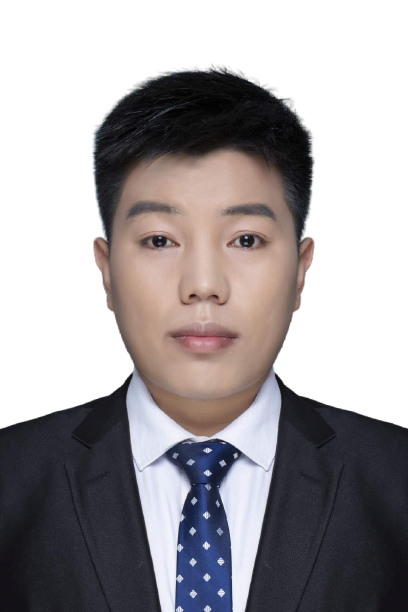}}]{Xiangzhu Meng}
received his B.S. degree from Anhui University, in 2015, and the Ph.D. degree in Computer Science and Technology from Dalian University of Technology, in 2021. Now, he is a postdoctoral researcher with the Center for Research on Intelligent Perception and Computing, Institute of Automation, Chinese Academy of Sciences, China. He regularly publishes papers in prestigious journals, including TNNLS, KBS, EAAI, INSC, APIN, etc. In addition, he serves as a reviewer for the conferences of ACM MM 2022 and MICAAI 2023. His research interests include multi-modal learning, and data mining.
\end{IEEEbiography}

\begin{IEEEbiography}[{\includegraphics[width=1in,height=1.25in,clip,keepaspectratio]{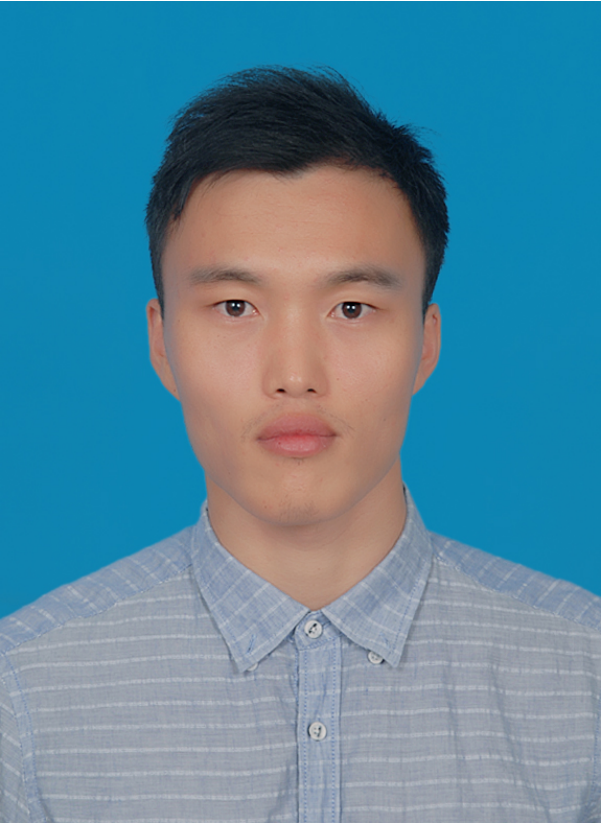}}]{Wei Wei} received the B.S. degree from the School of Mathematics and Information Science, Henan University, in 2012, and the Ph.D. degree in Institute of Systems Engineering from the Dalian University of Technology, in 2018. He is an associate professor of the Center for Energy, Environment \& Economy Research, College of Tourism Management, Zhengzhou University. His research interests include energy policy analysis, text mining, machine learning, and artificial intelligence.
\end{IEEEbiography}

\begin{IEEEbiography}[{\includegraphics[width=1in,height=1.25in,clip,keepaspectratio]{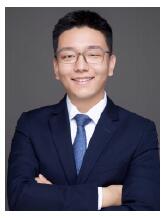}}]{Qiang Liu}
received his B.S. degree from is an Associate Professor in Center for Research on Intelligent Perception and Computing (CRIPAC), Institute of Automation, Chinese Academy of Sciences (CASIA). He received his PhD degree in pattern recognition from CASIA. Currently, his research interests include data mining, recommender systems, text mining, knowledge graph, and causal inference. He has published more than 30 papers in top-tier journals and conferences, such as ITKDE, AAAI, IJCAI, NeurIPS, WWW, SIGIR, CIKM and ICDM.
\end{IEEEbiography}

\begin{IEEEbiography}[{\includegraphics[width=1in,height=1.25in,clip,keepaspectratio]{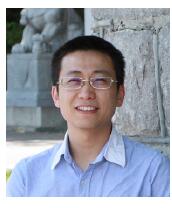}}]{Shu Wu}
received his B.S. degree from Hunan University, China, in 2004, M.S. degree from Xiamen University, China, in 2007, and Ph.D. degree from Department of Computer Science, University of Sherbrooke, Quebec, Canada, all in computer science. He is an Associate Professor with the Center for Research on Intelligent Perception and Computing (CRIPAC) at National Laboratory of Pattern Recognition (NLPR), Institute of Automation, Chinese Academy of Sciences (CASIA). He has published more than 70 papers in the areas of data mining and information retrieval in international journals and conferences, such as IEEE TKDE, IEEE THMS, AAAI, ICDM, SIGIR, and CIKM. His research interests include data mining and information retrieval.
\end{IEEEbiography}

\begin{IEEEbiography}[{\includegraphics[width=1in,height=1.25in,clip,keepaspectratio]{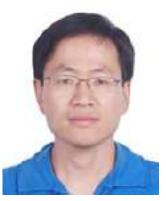}}]{Liang Wang}
received both the B.E. and M.S. degrees from Anhui University in 1997 and 2000, respectively, and the Ph.D. degree from the Institute of Automation, Chinese Academy of Sciences (CASIA) in 2004. Currently, he is a full professor of the Hundred Talents Program at the National Lab of Pattern Recognition, CASIA. His major research interests include machine learning, pattern recognition, and computer vision. He has widely published in highly ranked international journals such as IEEE TPAMI, IEEE TKDE and IEEE TIP, and leading international conferences such as CVPR, ICCV and ICDM. He is an IEEE and IAPR Fellow.
\end{IEEEbiography}

\end{document}